\documentclass[10pt,twocolumn,letterpaper]{article}

\usepackage{iccv}
\usepackage{times}
\usepackage{epsfig}
\usepackage{graphicx}
\usepackage{amsmath}
\usepackage{amssymb}
\usepackage{multicol}
\usepackage{multirow}
\usepackage{enumitem}
\usepackage{nccmath}
\usepackage{booktabs}
\usepackage{bm}
\usepackage{pifont}% http://ctan.org/pkg/pifont
\usepackage{pdfpages}
\newcommand{\xmark}{\ding{55}}%

\usepackage[breaklinks=true,bookmarks=false]{hyperref}

\iccvfinalcopy % *** Uncomment this line for the final submission

\def\iccvPaperID{2859} % *** Enter the ICCV Paper ID here

\begin{document}

%%%%%%%%% TITLE

\newcommand{\mymodelname}{LIGA-Stereo}
\title{\mymodelname: Learning LiDAR Geometry Aware Representations for Stereo-based 3D Detector}

\author{Xiaoyang Guo~~~~~
Shaoshuai Shi~~~~~
Xiaogang Wang~~~~~
Hongsheng Li\\
CUHK-SenseTime Joint Laboratory, The Chinese University of Hong Kong \\
{\tt\small \{xyguo, ssshi, xgwang, hsli\}@ee.cuhk.edu.hk}
}

\maketitle
\ificcvfinal\thispagestyle{empty}\fi

%%%%%%%%% ABSTRACT
\begin{abstract}
% Basic introduction of stereo-based 3D detection
Stereo-based 3D detection aims at detecting 3D objects from stereo images, which provides a low-cost solution for 3D perception. However, its performance is still inferior compared with LiDAR-based detection algorithms. To detect and localize accurate 3D bounding boxes, LiDAR-based detectors encode high-level representations from LiDAR point clouds, such as accurate object boundaries and surface normal directions. In contrast, high-level features learned by stereo-based detectors are easily affected by the erroneous depth estimation due to the limitation of stereo matching. To solve the problem, we propose \mymodelname~(\textbf{Li}DAR \textbf{G}eometry \textbf{A}ware Stereo Detector) to learn stereo-based 3D detectors under the guidance of high-level geometry-aware representations of LiDAR-based detection models. In addition, we found existing voxel-based stereo detectors failed to learn semantic features effectively from indirect 3D supervisions. We attach an auxiliary 2D detection head to provide direct 2D semantic supervisions. 
Experiment results show that the above two strategies improved the geometric and semantic representation capabilities. 
Compared with the state-of-the-art stereo detector, our method has improved the 3D detection performance of \textit{cars, pedestrians, cyclists} by {\it 10.44\%, 5.69\%, 5.97\%} mAP respectively on the official KITTI benchmark. The gap between stereo-based and LiDAR-based 3D detectors is further narrowed. The code is available at \url{https://xy-guo.github.io/liga/}.
\end{abstract}

%%%%%%%%%%%%%%%%%%%%%%%%%%%%%%%%%%%% Introduction %%%%%%%%%%%%%%%%%%%%%%%%%%%%%%%%%%%% 
\section{Introduction}
\label{sec:intro}
% intro 1. Basic overview of 3D detection and stereo-based 3D detection
% Recent progress of stereo-based deteciton and its problem
In recent years, LiDAR-based 3D detection~\cite{second,lang2019pointpillars,pointrcnn,pvrcnn,std,3dssd} has achieved increasing performance and stability in autonomous driving and robotics. However, the high cost of LiDAR sensors has limited its applications in low-cost products. Stereo matching~\cite{psmnet,groupwisestereo,gcnet,dispnet} is the most common depth sensing technique using only cameras. Compared with LiDAR sensors, stereo cameras are at a much lower cost and higher resolutions, which makes it a suitable alternative solution for 3D perception. 3D detection from stereo images aims at detecting objects using estimated depth maps~\cite{ocstereo,pseudolidar,pseudo++} or implicit 3D geometry representations~\cite{stereorcnn,dsgn,plume}. 
%from stereo image pairs. 
However, the performance of existing stereo-based 3D detection algorithms is still inferior compared with LiDAR-based algorithms.

\begin{figure}[t]
\begin{center}
    \includegraphics[width=\linewidth]{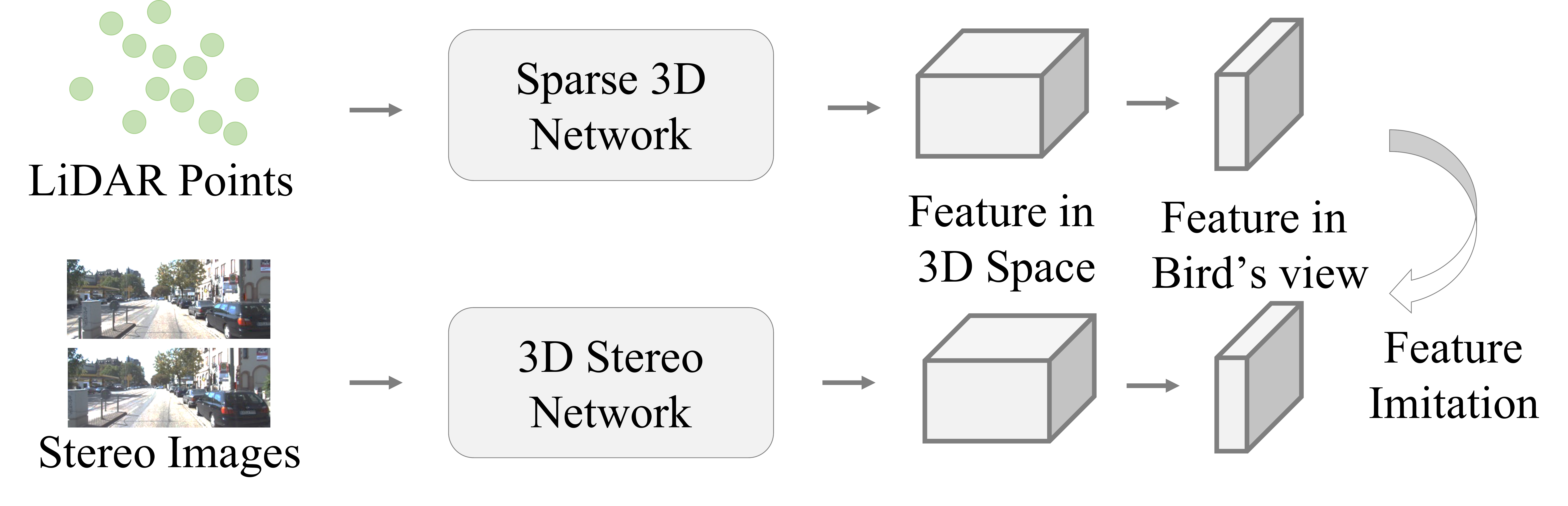}
\end{center}
\caption{Our method utilizes superior geometry-aware features from LiDAR-based 3D detection models to guide the training of stereo-based 3D detectors.}
\label{fig:framework_overview}
\end{figure}

% intro.2.1 advantages of LiDAR detection and its features; and our motivation
LiDAR-based detection algorithms take raw point cloud as inputs and then encode the 3D geometry information into intermediate and high-level feature representations. To detect and localize accurate 3D bounding boxes, the model must learn robust local features about object boundaries and surface normal directions, which are essential for predicting accurate bounding box size and orientation. The features learned by LiDAR-based detectors provide robust high-level summarization of accurate 3D geometry structures. In comparison, due to the limitation of stereo matching, the inaccurately estimated depth or implicit 3D representation have difficulties in encoding accurate 3D geometry of objects, especially for distant ones. In addition, the target box supervisions only provide object-level supervisions (location, size, and orientation).

% intro.2.2 Briefly introduce our idea
This inspires us to utilize superior LiDAR detection models to guide the training of stereo detection model via imitating the geometry-aware representations encoded by the LiDAR model.
Comparing with traditional knowledge distillation~\cite{distilling_hinton2015} for recognition tasks, we did not take the final erroneous classification and regression predictions from the LiDAR model as ``soft'' targets, which we found benefits little for training stereo detection networks. The erroneous regression targets would constrain the upper-bound accuracy of bounding box regression. Instead, we force our model to align intermediate features with those of LiDAR models, which encode high-level geometry representations of the scene. The features from LiDAR models could provide powerful and discriminative high-level geometry-aware features, such as surface normal directions and boundary locations. On the other hand, the LiDAR features can provide extra regularization to alleviate the over-fitting problem caused by erroneous stereo predictions.

% intro.3 other contributions
Besides learning better geometry features, we further explore how to learn better semantic features for boosting the 3D detection performance. Instead of learning semantic features from indirect 3D supervisions, we propose to attach an auxiliary multi-scale 2D detection head on the 2D semantic features, which could directly guide the learning of 2D semantic features. Our baseline model, Deep Stereo Geometry Network (DSGN)~\cite{dsgn}, failed to benefit from extra semantic features effectively according to their ablation studies. We argue that the network provides erroneous semantic supervisions from indirect 3D supervisions because of depth estimation errors, while our proposed direct guidance could greatly benefit 3D detection performance from better learning of 2D semantic features. Experiment results show that the performance is further improved, especially for classes with few samples like \textit{cyclist}.

% contributions
The contributions can be summarized as follows. 1) We propose to utilize features from superior LiDAR-based detection models to guide the training of stereo-based 3D detection model. LiDAR features encode compact 3D geometry representations of the scene to guide and regularize the stereo features. 2) By attaching an auxiliary 2D detection head to provide direct 2D supervisions, our model significantly improves the learning efficiency for semantic features, which further improves the recall rate especially for rare categories. 4) On the official KITTI 3D detection benchmark, our proposed method surpasses state-of-the-art models by {\it 10.44\%}, {\it 5.69\%}, and {5.97\%} mAP on the \textit{car}, \textit{pedestrian} and \textit{cyclist} classes respectively.

\begin{figure*}[t]
\begin{center}
    \includegraphics[width=\linewidth]{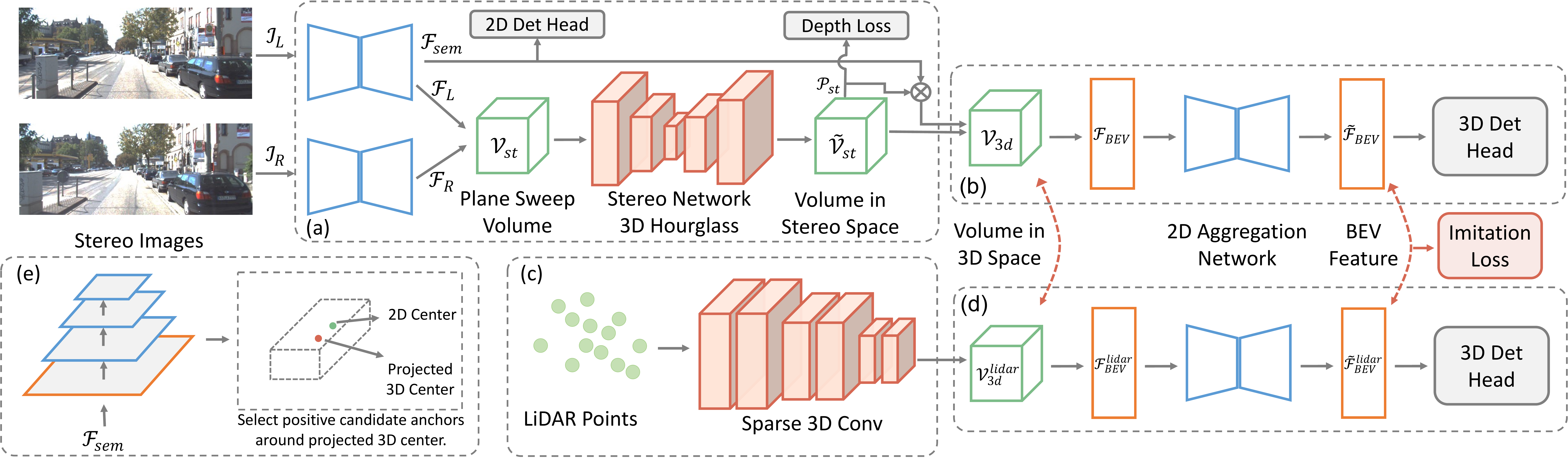}
\end{center}
\caption{The framework of our stereo-based 3D detection algorithm. (a) The stereo matching network which takes a stereo image pair as inputs and outputs a feature volume in 3D space. (b) and (d) share the same structure, which detects 3D objects from dense / sparse 3D feature volumes, respectively. (c) The teacher LiDAR-based detection network~\cite{second}, which processes raw LiDAR point clouds using sparse 3D convolution layers. (e) Our auxiliary 2D detection head for direct semantic supervisions.}
\label{fig:framework}
\end{figure*}

%%%%%%%%%%%%%%%%%%%%%%%%%%%%%%%%%%%% Related Work %%%%%%%%%%%%%%%%%%%%%%%%%%%%%%%%%%%% 
\section{Related Work}
\noindent\textbf{Stereo Matching.}
Mayer \etal~\cite{dispnet} proposed the first deep stereo algorithm DispNet, which regresses disparity map from feature-based correlation cost volume. DispNet is then extended using multi-stage refinement~\cite{crl,iresnet} and auxiliary semantic features~\cite{edgestereo,segstereo}. 
State-of-the-art stereo models construct feature-based cost volume by concatenating left-right 2D features for all disparity candidates and then apply 3D aggregation network to predict the disparity probability distribution~\cite{gcnet,psmnet,groupwisestereo,unimodal}. State-of-the-art stereo detection networks~\cite{pseudolidar,pseudo++,dsgn,plume} also estimate depth by similar network structures. Zhang \etal~\cite{ganet} proposed novel cost volume aggregation strategies to improve the computational efficiency. Yin \etal~\cite{hd3} accelerated stereo matching by hierarchically estimating local disparity distributions at each scale and composing them together to form the final match density. Xu \etal~\cite{aanet} proposed a sparse points based intra-scale cost aggregation to alleviate the edge-fattening issue. Cheng \etal~\cite{nasstereo} employed neural architecture search (NAS) to automatically search the optimal network structure for stereo matching.

\noindent\textbf{LiDAR-based 3D Detection. }
By leveraging the more accurate depth information captured by LiDAR sensors, 3D detection approaches~\cite{zhou2018voxelnet,qi2018frustum,yang2018pixor,pointrcnn,lang2019pointpillars,shi2020points, pvrcnn} with LiDAR point clouds generally achieve better performance than image-based approaches.
To learn effective features from irregular and sparse point clouds, most existing approaches adopt the voxelization operation to transfer point clouds to regular grids, where the 3D space is first divided into regular 3D voxels~\cite{zhou2018voxelnet,shi2020points} or bird-view 2D grids~\cite{yang2018pixor,lang2019pointpillars} to be processed by convolutions for detection head. Yan \etal~\cite{second} proposed to utilize sparse convolutions~\cite{3DSemanticSegmentationWithSubmanifoldSparseConvNet} for efficient feature learning from sparse voxels.
Du \etal~\cite{du2020associate} presented a feature imitation strategy to learn better perceptual features from synthesized conceptual scenes. Inspired by them, we propose to imitate the much informative feature maps from LiDAR models for better guidance beyond 3D box annotations. 

\noindent\textbf{Stereo-based 3D Detection. }
Stereo-based 3D detection algorithms can be roughly divided into three types: 

1) \textit{2D-based methods}~\cite{3dop-pami,mlfstereo,stereorcnn,ocstereo,zoomnet,disprcnn,ida3d} first detect 2D bounding box proposals and then regress instance-wise 3D boxes. Stereo-RCNN~\cite{stereorcnn} extended Faster R-CNN~\cite{fasterrcnn} for stereo-inputs to associate left and right images. Disp R-CNN~\cite{disprcnn} and ZoomNet~\cite{zoomnet} incorporated extra instance segmentation mask and part location map to improve detection quality. However, the final performance is limited by the recall of 2D detection algorithms, and 3D geometry information is not fully utilized.

2) \textit{Pseudo-LiDAR}~\cite{pseudolidar,pseudo++,pseudo_e2e,cgstereo,cdn} based 3D detection first estimate depth maps and then detect 3D bounding boxes using existing LiDAR-based algorithms. Pseudo-LiDAR++~\cite{pseudo++} adapted stereo cost volume to depth cost volume for direct depth estimation. Qian \etal~\cite{pseudo_e2e} made Pseudo-LiDAR pipeline end-to-end trainable. However, these models only take geometry information into consideration, which is lack of complementary semantic features. 
% Garg \etal~\cite{} proposed to predict a real-value offset to solve the inaccuracy problem caused by pre-defined discrete disparity values. 

3) Volume-based methods construct 3D anchor space \cite{triangulation} or detect from 3D stereo volume \cite{dsgn,plume}. DSGN~\cite{dsgn} directly constructs differentiable volumetric representation, which encodes implicit 3D geometry structure of the scene, for one-stage stereo-based 3D detection. PLUME~\cite{plume} directly constructs geometry volume in 3D space for acceleration. Our work takes DSGN~\cite{dsgn} as the baseline model. We solve several key problems existed in \cite{dsgn} and surpass state-of-the-art models by a large margin.

\noindent\textbf{Knowledge Distillation.}
Distillation is first proposed by Hinton \etal~\cite{distilling_hinton2015} for model compression by supervising student networks with ``softened labels'' from predictions of large teacher networks. Going further from ``softened labels'', knowledge from intermediate layers provide richer information from the teacher~\cite{fitnet,distill_feature_heo2019comprehensive,yim2017gift,distill_featurehuang2017like,distill_feature_kim2018paraphrasing}. Recently, knowledge distillation has been successfully applied to detection~\cite{distill_object_det_chen2017,distilling_fine_grained_detectors_wang,chen2019_learning_ped_detector_with_distillation,xu2019_training_binary_obj_detector,feliz2020_squeezed_6dot_detection_distillation} and semantic segmentation~\cite{mullapudi2019_online_distillation_for_video_inference,liu2019_structured_distillation_for_sem_seg,he2019_knowledge_adapt_for_efficient_sem_seg,chen2018_reality_oriented_adapt_for_sem_seg}. The knowledge can also be transferred across modalities~\cite{crossmodeldistill_gupta2016,cross_modal_distill_for_lip_reading,aytar2016soundnet,knowledgeaspriors}. However, the feature imitation from LiDAR-based to stereo-based 3D detectors is first explored in this paper.

% Cross-modal distilling is designed to distill features from a superior modality to improve the training of another weak modality. Different from multi-modal learning, the final network will not take data from the superior modality as inputs. Cross-modal distillation has been proved to be useful to improve depth-map features~\cite{crossmodeldistill_gupta2016} or raw audio representations~\cite{aytar2016soundnet} with visual supervisions. 
% transfer learning across different modalities~\cite{crossmodeldistill_gupta2016,knowledgeaspriors,cross_modal_distill_for_lip_reading}. 
% Cross-modal knowledge distillation~\cite{crossmodeldistill_gupta2016,knowledgeaspriors,cross_modal_distill_for_lip_reading} extended knowledge distillation to transfer knowledge across models with different modality inputs. The teacher model is usually trained with superior modality to supervised another model trained with weak modality, for example, from a image-base recognition model to a depth-based recognition model~\cite{crossmodeldistill_gupta2016} or an audio model~\cite{aytar2016soundnet}. 

%%%%%%%%%%%%%%%%%%%%% Our Approach %%%%%%%%%%%%%%%%%%%%%%%%%%%%%%%%%%%% 
\section{Our Approach}
In this work, to improve the performance of our model, we developed two strategies to learn better geometric and semantic features respectively. Due to the limitation of stereo matching, stereo-based detectors are fragile to the erroneous depth estimation, especially for low-texture surfaces, blurry boundaries and occlusion areas. In comparison, the features learned by LiDAR-based detectors provide robust high-level geometry-aware representations (accurate boundaries and surface normal directions). To minimize the gap between LiDAR-based and stereo-based detectors, we propose to utilize LiDAR models to guide the training of stereo-based detectors for better geometric supervision. In addition, we employ auxiliary 2D semantic supervisions to improve the learning efficiency for semantic features. 

In the following section, we will first revisit the baseline model, Deep Stereo Geometry Network (DSGN)~\cite{dsgn}, in Sec.~\ref{sec:review-dsgn} to make the paper self-contained. In Sec.~\ref{sec:distilling-lidar}, we describe the proposed LiDAR imitation strategy for encoding better geometry representations. In Sec.~\ref{sec:aux-2d}, we discuss why the baseline model is of low efficiency for learning semantic features and propose the corresponding solution. The training losses are specified in Sec.~\ref{sec:losses}. 

\subsection{Revisit of Deep Stereo Geometry Network}
\label{sec:review-dsgn}
In this paper, we utilize Deep Stereo Geometry Network (DSGN)~\cite{dsgn} as our baseline model, which directly detects objects using implicit volumetric representations. 

\noindent\textbf{Volume in Stereo Space. } Given a left-right image pair $(\mathcal{I}_L, \mathcal{I}_R)$ and their features $(\mathcal{F}_L, \mathcal{F}_R)$, a plane-sweep volume $\mathcal{V}_{st}$ is constructed by concatenating left features and corresponding right features for every candidate depth level,
\begin{equation}
    \mathcal{V}_{st}(u,v,w) = \text{concat}\left[\mathcal{F}_L(u, v), \mathcal{F}_R(u - \frac{fL}{d(w)s_\mathcal{F}}, v)\right],
\label{equ:v_st}
\end{equation}
in which $(u,v)$ is the current pixel coordinate . 
$w{=}$$0,1,\cdots$ is the candidate depth index, and $d(w)\texttt{=} w\cdot v_d{+}z_{min}$ is the function to calculate its corresponding depth, where $v_d$ is the depth interval, and $z_{min}$ is the minimum depth of the detection area.
$f$, $L$, and $s_\mathcal{F}$ are the camera focal length, the baseline of the stereo camera pair, and the stride of the feature map, respectively. After filtering $\mathcal{V}_{st}$ with a 3D cost volume aggregation network, we obtain an aggregated stereo volume $\mathcal{\tilde{V}}_{st}$ and a depth distribution volume $\mathcal{P}_{st}$. $\mathcal{P}_{st}(u, v, :)$ represents the depth probability distribution of pixel $(u, v)$ over all discrete depth levels $d(w)$.

\noindent\textbf{Volume in 3D Space. } In order to convert the feature volume from stereo space to normal 3D space, the 3D detection area is divided into small voxels of the same size. For each voxel, we project its center $(x, y, z)$ back into the stereo space using the feature intrinsics $\mathbf{K_\mathcal{F}}$ to obtain its reprojected pixel coordinate $(u, v)$ and depth index $d^{-1}(z){=}{(z{-}z_{min})}/{v_d}$. The volume in the 3D space is then defined as the concatenation of resampled stereo volume and semantic features masked by depth probability,
\begin{equation}
\begin{split}
    \mathcal{V}_{3d}(x,y,z) = \text{con}&\text{cat}\Big[
    \mathcal{V}_{st}\left(u, v, d^{-1}(z)\right), \\
    & \mathcal{F}_{sem}(u,v) \cdot \mathcal{P}_{st}\left(u, v, d^{-1}(z)\right) \Big],
\end{split}
\label{equ:v_3d}
\end{equation}
in which $\mathcal{V}_{st}$ and $\mathcal{F}_{sem}$ provide geometric and semantic features respectively. Note that we ignore the trilinear and bilinear resampling operators for simplicity.

\noindent\textbf{Feature in BEV Space and Detection Head. } The 3D volume $\mathcal{V}_{3d}$ is then rearranged into 2D bird's eye view (BEV) feature $\mathcal{F}_{BEV}$ by merging the channel dimension and the height dimension as \cite{dsgn}. A 2D aggregation network and detection heads are attached to $\mathcal{F}_{BEV}$ to generate the aggregated BEV feature $\mathcal{\tilde{F}}_{BEV}$ and predict the final 3D bounding boxes, respectively. The training loss is divided into two parts, depth regression loss and 3D detection loss, 
\begin{equation}
    \mathcal{L}_{bsl}=\mathcal{L}_{depth}+\mathcal{L}_{det}.
\label{equ:loss_bsl}
\end{equation}
We have made several modifications to the above baseline to improve both performance and efficiency. Please see details in Sec.~\ref{sec:losses} and Sec.~1 of the supplementary materials.

\subsection{Learning LiDAR Geometry-aware Representations}
\label{sec:distilling-lidar}

LiDAR-based detection models~\cite{zhou2018voxelnet,qi2018frustum,yang2018pixor,pointrcnn,lang2019pointpillars,shi2020points, pvrcnn} take raw point clouds as inputs, which are then encoded into high-level features (such as accurate boundaries and surface normal directions) for accurate bounding box localization. 
For stereo-based models, pure depth loss and detection loss could not well make the model learn such features for occlusion areas, non-textured areas and distant objects due to erroneous depth representations. 

Inspired by the above observation, we propose to guide the training of stereo-based detectors using the high-level geometry-aware features from LiDAR models. In this paper, we utilize SECOND~\cite{second} as our LiDAR ``teacher''. To make the features as consistent as possible between two models, we employ 2D aggregation network and detection head of the same structure, as shown in Fig.~\ref{fig:framework} (see details in the supplementary materials).

Due to the huge differences between the backbone of the LiDAR model and the stereo model, we only focus on how to perform feature imitation for high-level features $\mathbb{F}_{im}{\subseteq}\{\mathcal{V}_{3d}$, $\mathcal{F}_{BEV}$, $\mathcal{\tilde{F}}_{BEV}\}$, which are of the same shape between the two models. 
Our imitation loss aims at minimizing the feature distances between stereo feature $\mathcal{F}_{im}{\in}\mathbb{F}_{im}$ and its corresponding LiDAR feature $\mathcal{F}_{im}^{lidar}$,
\begin{equation}
    \mathcal{L}_{im}{=}\sum_{\mathcal{F}_{im}{\in}\mathbb{F}_{im}}\frac{1}{N_{pos}} \left\| M_{fg} M_{sp} \left( g(\mathcal{F}_{im}) {-} \frac{\mathcal{F}_{im}^{lidar}}{\mathbb{E}\left[|\mathcal{F}_{im}^{lidar}|\right]} \right) \right\|_2^2,
\label{equ:distill}
\end{equation}
where $g$ is a single convolution layer with kernel size 1 followed by an optional ReLU layer (depending on whether ReLU is applied to $\mathcal{F}_{im}^{lidar}$). $M_{fg}$ is the foreground mask, which is 1 inside any ground-truth object box, and 0 otherwise. $M_{sp}$ is the sparse mask of $\mathcal{F}_{im}^{lidar}$, which is 1 for non-empty indices and 1 otherwise. $N_{pos}=\sum{M_{fg}M_{sp}}$ is the normalization factor. Note that the LiDAR feature $\mathcal{F}_{im}^{lidar}$ is normalized by the channel-wise expectation of non-zero absolute values of $\mathcal{F}_{im}^{lidar}$ to make sure the scale stability of the L2 loss.

Experiment results show that the detection performance benefit little from imitating whole feature maps, thus $M_{fg}$ is essential to make the imitation loss focus on foreground objects. We found that $\mathcal{{F}}_{BEV}$ and $\mathcal{{V}}_{3d}$ provide the most effective supervisions for training our stereo detection network. Please check the ablation studies in Sec.~\ref{sec:ablation-distill}. 

\begin{figure}[tb]
\begin{center}
    \includegraphics[width=\linewidth]{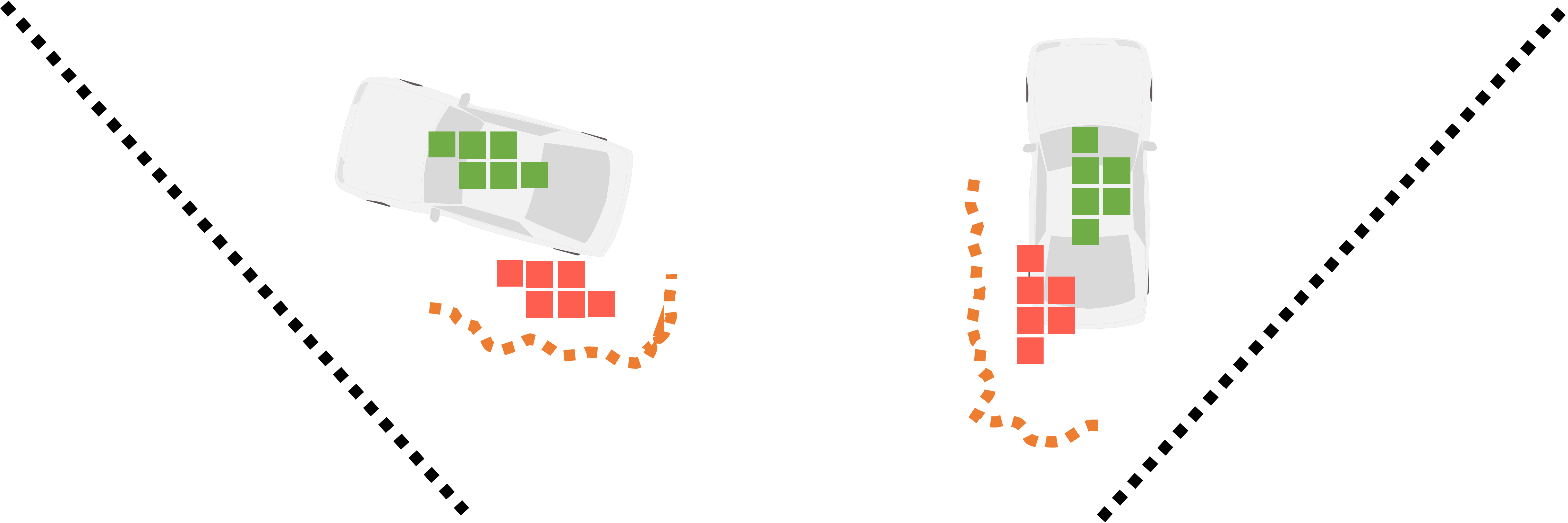}
\end{center}
\caption{Illustration of incorrect semantic supervisions caused by depth estimation errors (drawn in brid's eye view). Green and red squares represent positive and negative anchors, and orange dash lines represent estimated surface with large errors. Values of $\mathcal{P}_{st}$ are large around estimated surfaces, thus semantic features of foreground objects are falsely resampled around the wrong surfaces. Therefore, the supervisions for semantic features are inaccurate.} 
\label{fig:semantic_errors}
\end{figure}

%%%%%% Table for val set %%%%%%%%% 
\begin{table*}[tb]\footnotesize
\begin{center}
\begin{tabular}{c|c|ccc|ccc|ccc|ccc}
\hline
% header
\multirow{2}{*}{Sensor} & \multirow{2}{*}{Method} & 
\multicolumn{3}{c|}{\textbf{AP$_{\text{3D}}$ (IoU=0.7)}} & 
\multicolumn{3}{c|}{AP$_{\text{BEV}}$ (IoU=0.7)} &
\multicolumn{3}{c|}{AP$_{\text{3D}}$ (IoU=0.5)} & 
\multicolumn{3}{c}{AP$_{\text{BEV}}$ (IoU=0.5)} \\
\cline{3-14}
& & Easy & \textbf{Mod} & Hard & Easy & Mod & Hard & Easy & Mod & Hard & Easy & Mod & Hard \\
\hline
% LiDAR results
\multirow{2}{*}{LiDAR}
& MV3D (LiDAR) \cite{MV3D} & 71.29 & 62.68 & 56.56 & 86.55 & 78.10 & 76.67 & -- & -- & -- & -- & -- & --\\
& PL++: P-RCNN + SL* \cite{pseudo++} & 75.1 & 63.8 & 57.4 & 88.2 & 76.9 & 73.4 & -- & -- & -- & -- & -- & -- \\ & SECOND \cite{second} & 87.43 & 76.48 & 69.10 & 89.96 & 87.07 & 79.66 & --  & --  & --  & --  & --  & -- \\
& Point-RCNN \cite{pointrcnn} & 88.88 & 78.63 & 77.38 & --  & --  & --  & --  & --  & --  & --  &--   &--  \\
& \textit{SECOND (our teacher)} & 88.82 & 78.57 & 77.40 & 89.93 & 87.75 & 86.67 & 98.12 & 90.17 & 89.64 & 98.16 & 90.20 & 89.71 \\
\hline
% Stereo results
\multirow{10}{*}{Stereo}
& 3DOP \cite{3dop} & 6.55 & 5.07 & 4.10 & 12.63 & 9.49 & 7.59 & 46.04 & 34.63 & 30.09 & 55.04 & 41.25 & 34.55 \\
& TLNet \cite{triangulation} & 18.15 & 14.26 & 13.72 & 29.22 & 21.88 & 18.83 & 59.51 & 43.71 & 37.99 & 62.46 & 45.99 & 41.92\\
& Stereo-RCNN \cite{stereorcnn} & 54.11 & 36.69 & 31.07 & 68.50 & 48.30 & 41.47 & 85.84 & 66.28 & 57.24 & 87.13 & 74.11 & 58.93 \\
& PL: F-PointNet \cite{pseudolidar} & 59.4 & 39.8 & 33.5 & 72.8 & 51.8 & 33.5 & 89.5 & 75.5 & 66.3 & 89.8 & 77.6 & 68.2  \\
& PL++: P-RCNN \cite{pseudo++} & 67.9 & 50.1 & 45.3 & 82.0 & 64.0 & 57.3 & 89.7 & 78.6 & 75.1 & 89.8 & 83.8 & 77.5  \\
& OC-Stereo \cite{ocstereo} & 64.07 & 48.34 & 40.39 & 77.66 & 65.95 & 51.20 & 89.65 &  80.03 & 70.34 & 90.01 & 80.63 & 71.06 \\
& Disp-RCNN \cite{disprcnn} & 64.29 & 47.73 & 40.11 & 77.63 & 64.38 & 50.68 & 90.47 & 79.76 & 69.71 & 90.67 & 80.45 & 71.03 \\
& DSGN \cite{dsgn} & 72.31 & 54.27 & 47.71 & 83.24 & 63.91 & 57.83 & -- & -- & -- & -- & -- & -- \\
& CG-Stereo \cite{cgstereo} & 76.17 & 57.82 & 54.63 & 87.31 & 68.69 & 65.80 & 90.58 & 87.01 & 79.76 & 97.04 & 88.58 & 80.34 \\
& PLUME-Large \cite{plume}  & -- & -- & -- & 84.7 & 71.1 & 65.1  & -- & -- & -- & 91.3 & 86.6 & 81.6 \\
& \textbf{\mymodelname~(Ours)} & \textbf{84.92} & \textbf{67.06} & \textbf{63.80} & \textbf{89.35} & \textbf{77.26} & \textbf{69.05} & \textbf{97.06} & \textbf{89.97} & \textbf{87.94} & \textbf{97.22} & \textbf{90.27} & \textbf{88.36} \\
\hline
\end{tabular}
\end{center}
\caption{Car detection results on the KITTI \textit{validation} set. The results are evaluated using the original KITTI metric with 11 recall values for fair comparison. If not specified, all results in other tables are evaluated using 40 recall positions. * utilizes 4-beam LiDAR to refine stereo depth estimation. }
\label{table:val-results-car}
\end{table*}

%%%%%% Table for car test set %%%%%%%%% 
\begin{table*}[tbh]\footnotesize
\begin{center}
\begin{tabular}{c|c|ccc|ccc|ccc}
\hline
% header
\multirow{2}{*}{Sensor} & \multirow{2}{*}{Method} & 
\multicolumn{3}{c|}{ \textbf{Car AP$_{\text{3D}}$} } & 
\multicolumn{3}{c|}{Car AP$_{\text{BEV}}$ } & 
\multicolumn{3}{c}{Car AP$_{\text{2D}}$ } \\ 
\cline{3-11}
& & Easy & \textbf{Moderate} & Hard & Easy & Moderate & Hard & Easy & Moderate & Hard \\ 
\hline
% LiDAR results
\multirow{5}{*}{LiDAR}
 & MV3D \cite{MV3D} & 74.97 & 63.63 & 54.00 & 86.62 & 78.93 & 69.80 & -- & -- & -- \\
%  & F-PointNet \cite{fpointnet} & 82.19 & 69.79 & 60.59 & 91.17 & 84.67 & 74.77 & -- & -- & -- \\
 & SECOND \cite{second} & 83.34 & 72.55 & 65.82 & 89.39 & 83.77 & 78.59 & -- & -- & -- \\
 & PointPillars \cite{lang2019pointpillars} & 82.58 & 74.31 & 68.99 & 90.07 & 86.56 & 82.81 & 94.00 & 91.19 & 88.17 \\
 & Point-RCNN \cite{pointrcnn} & 86.96 & 75.64 & 70.70 & 92.13 & 87.39 & 82.72 & 95.92 & 91.90 & 87.11 \\
 & PV-RCNN \cite{pvrcnn} & 90.25 & 81.43 & 76.82 & 94.98 & 90.65 & 86.14 & 98.17 & 94.70 & 92.04 \\  
 & PL++ (SDN+GDC)* \cite{pseudo++} & 68.38 & 54.88 & 49.16 & 84.61 & 73.80 & 65.59 & 94.95 & 85.15 & 77.78 \\
\hline
% Stereo results
\multirow{10}{*}{Stereo}
% & 3DOP \cite{3dop} & -- & -- & -- & -- & -- & -- & 93.04 & 88.64 & 79.10 \\
& Stereo R-CNN \cite{stereorcnn} & 47.58 & 30.23 & 23.72 & 61.92 & 41.31 & 33.42 & 93.98 & 85.98 & 71.25 \\
& PL: AVOD \cite{pseudolidar} & 54.53 & 34.05 & 28.25 & 67.30 & 45.00 & 38.40 & 85.40 & 67.79 & 58.50 \\
& PL++: P-RCNN \cite{pseudo++} & 61.11 & 42.43 & 36.99 & 78.31 & 58.01 & 51.25 & 94.46 & 82.90 & 75.45 \\
& ZoomNet \cite{zoomnet} & 55.98 & 38.64 & 30.97 & 72.94 & 54.91 & 44.14 & 94.22 & 83.92 & 69.00 \\
& OC-Stereo \cite{ocstereo} & 55.15 & 37.60 & 30.25 & 68.89 & 51.47 & 42.97 & 87.39 & 74.60 & 62.56 \\
& Disp-RCNN \cite{disprcnn} & 68.21 & 45.78 & 37.73 & 79.76 & 58.62 & 47.73 & 93.45 & 82.64 & 70.45 \\
& DSGN \cite{dsgn} & 73.50 & 52.18 & 45.14 & 82.90 & 65.05 & 56.60 & 95.53 & 86.43 & 78.75 \\
& CG-Stereo \cite{cgstereo} & 74.39 & 53.58 & 46.50 & 85.29 & 66.44 & 58.95 & 96.31 & 90.38 & 82.80 \\
& CDN \cite{cdn} & 74.52 & 54.22 & 46.36 & 83.32 & 66.24 & 57.65 & 95.85 & 87.19 & 79.43 \\
& PLUME-Middle \cite{plume} & -- & -- & -- & 83.0 & 66.3 & 56.7 & -- & -- & -- \\
& \textbf{\mymodelname~(Ours)} & {\bf 81.39} & {\bf 64.66} & {\bf 57.22} & {\bf 88.15} & {\bf 76.78} & {\bf 67.40} & {\bf 96.43} & {\bf 93.82} & {\bf 86.19} \\
\hline
\end{tabular}
\end{center}
\caption{Car detection results on the KITTI \textit{test} set (official KITTI leaderboard). * utilizes 4-beam LiDAR to refine stereo depth estimation. }
\label{table:test-results-car}
\end{table*}

%%%%%% Table for ped/cyclist test set %%%%%%%%% 
\begin{table*}[tbh]\footnotesize
\begin{center}
\begin{tabular}{c|ccc|ccc|ccc|ccc}
\hline
% header
\multirow{2}{*}{Method} & 
\multicolumn{3}{c|}{\textbf{Pedestrian AP$_{\text{3D}}$}} & 
\multicolumn{3}{c|}{Pedestrian AP$_{\text{BEV}}$} & 
\multicolumn{3}{c|}{\textbf{Cyclist  AP$_{\text{3D}}$}} & 
\multicolumn{3}{c}{Cyclist AP$_{\text{BEV}}$} \\
\cline{2-13}
& 
Easy & \textbf{Moderate} & Hard & Easy & Moderate & Hard & 
Easy & \textbf{Moderate} & Hard & Easy & Moderate & Hard \\ 
\hline
% Stereo results
OC-Stereo \cite{ocstereo} & 24.48 & 17.58 & 15.60 & 29.79 & 20.80 & 18.62 & 29.40 & 16.63 & 14.72 & 32.47 & 19.23 & 17.11\\
DSGN \cite{dsgn} & 20.53 & 15.55 & 14.15 & 26.61 & 20.75 & 18.86 & 27.76 & 18.17 & 16.21 & 31.23 & 21.04 & 18.93 \\
Disp-RCNN \cite{disprcnn} & 37.12 & 25.80 & 22.04 & 40.21 & 28.34 & 24.46 & 40.05 & 24.40 & 21.12 & 44.19 & 27.04 & 23.58  \\
CG-Stereo \cite{cgstereo} & 33.22 & 24.31 & 20.95 & 39.24 & 29.56 & 25.87 & 47.40 & 30.89 & 27.73 & 55.33 & 36.25 & 32.17 \\
\hline

\mymodelname~(Ours) & \textbf{40.46} & \textbf{30.00} & \textbf{27.07} & \textbf{44.71} & \textbf{34.13} & \textbf{30.42} & \textbf{54.44} & \textbf{36.86} & \textbf{32.06} & \textbf{58.95} & \textbf{40.60} & \textbf{35.27} \\
\hline
\end{tabular}
\end{center}
\caption{Pedestrian and cyclist detection results on KITTI \textit{test} set (official KITTI leaderboard).}
\label{table:test-results-ped-cyc}
\end{table*}

\subsection{Improving Semantic Features by Direct 2D Supervisions}
\label{sec:aux-2d}
From Eq.~\eqref{equ:v_3d} we can see before resampling semantic feature into 3D space, it is first multiplied by the depth probability from $\mathcal{P}_{st}$ (which can be seen as the estimation of 3D occupancy mask. Please see how we supervise $\mathcal{P}_{st}$ in Eq.~\eqref{equ:loss_depth}). In this way, semantic features are only resampled near the estimated surface (orange dashed lines in Fig.~\ref{fig:semantic_errors}).
However, when there exist large errors in the estimated depth values, the semantic features will be resampled to the wrong positions as illustrated in Fig.~\ref{fig:semantic_errors}. As a result, the resampled semantic features  deviate from the ground-truth positions and are then assigned with negative anchors (red squares in Fig.~\ref{fig:semantic_errors}), and there is no resampled semantic feature around the positive anchors (green squares in Fig.~\ref{fig:semantic_errors}). Therefore, the supervision signals for the semantic features are incorrect in this case, which causes the low learning efficiency of semantic features.

To solve the problem, we add an auxiliary 2D detection head to provide direct supervisions for learning semantic features. Instead of creating feature pyramids (FPN) using multi-level features like \cite{fpn,atss}, we take only a single feature map $\mathcal{F}_{sem}$ to construct feature pyramids as shown in Fig.~\ref{fig:framework}(e), which forms an information ``bottleneck'' to enforce all the semantic features to be encoded into $\mathcal{F}_{sem}$. Five consecutive convolution layers with a stride of 2 are attached to $\mathcal{F}_{sem}$ to construct a multi-level feature pyramid, which are then connected to an ATSS~\cite{atss} head for 2D detection. Since we find that the dilated convolutions and spatial pyramid pooling (SPP)~\cite{spp} have already produced highly semantic features with large receptive fields, we ignore the top-down path of FPN for simplicity. Please see detailed network structures in Sec. 1 of the supplementary materials.

To ensure the semantic alignment between 2D and 3D features, 2D detection head should predict high scores around re-projected 3D object centers. 
We made a small modification to the positive sample assignment algorithm of ATSS~\cite{atss}.
For each ground-truth bounding box $g$, we select $k$ candidate anchors from each scale if their centers are closest to the re-projected 3D object centers, instead of 2D bounding box centers as in~\cite{atss}. The candidate anchors are then filtered by the dynamic IoU threshold as in ATSS~\cite{atss} to assign the final positive samples.

\subsection{Modifications to Baseline and Training Losses}
\label{sec:losses}

Note that we made several important modifications to DSGN to obtain a faster and more robust baseline model. 
1) Decrease the number of channels and layers to reduce memory consumption and computation cost. 
2) Utilize SECOND~\cite{second} detection head for 3D detection. 
3) Replace the smooth-L1 depth regression loss with the uni-modal depth loss~\cite{hd3} in Eq.~\eqref{equ:loss_depth} based on Kullback-Leibler divergence. 
4) Use a combination of L1 loss and auxiliary rotated 3D IoU loss~\cite{3diou} for better bounding box regression. 
5) Attach a small U-Net to the 2D backbone to encode full-resolution feature maps for stereo volume construction.
Please check Sec.~1 \& 2.1 of the supplementary material for detailed network structure and performance analysis for these modifications. 

The new overall loss of our model is formulated as
\begin{equation}
\begin{split}
    &\mathcal{L}
    =\mathcal{L}_{depth}+\mathcal{L}_{det}+\lambda_{im}\mathcal{L}_{im}+\lambda_{2d}\mathcal{L}_{2d}, \\
    &\mathcal{L}_{det}=\mathcal{L}_{cls}{+}\lambda_{reg}^{L1}\mathcal{L}_{reg}^{L1}{+}\lambda_{reg}^{IoU}\mathcal{L}_{reg}^{IoU}{+}\lambda_{dir}\mathcal{L}_{dir},
\label{equ:loss_final}
\end{split}
\end{equation}
where $\mathcal{L}_{cls}$, $\mathcal{L}_{reg}^{L1}$, and $\mathcal{L}_{dir}$ are the same classification loss, box regression loss, and direction classification loss as in SECOND~\cite{second}. $\mathcal{L}_{reg}^{IoU}$ is the average rotated IoU loss~\cite{3diou} between 3D box predictions and ground-truth bounding boxes. The uni-modal depth loss $\mathcal{L}_{depth}$ is formulated as
\begin{equation}
\begin{split}
    &\mathcal{L}_{depth}=\frac{1}{N_{gt}}\sum_{u,v}\sum_{w} \bigg[ \\
    & -\max\left(1-\frac{|d^* - d(w)|}{v_{d}}, 0\right) \log{\mathcal{P}_{st}(u, v, w)}
    \bigg],
\end{split}
\label{equ:loss_depth}
\end{equation}
in which $d^*$ is the ground-truth depth. Note that the loss is only applied to the pixels $(u,v)$ with valid LiDAR depths, and $N_{gt}$ denotes the number of valid pixels. Compared with the original L1 loss, the loss in Eq.~\eqref{equ:loss_depth} provides more concentrated supervisions to the target distribution. 

%%%%%%%%%%%%%%%%%%%%%%%%%%%%%%%%%%%% Experiment results %%%%%%%%%%%%%%%%%%%%%%%%%%%%%%%%%%%% 

%%%%%% Table for main ablation studies %%%%%%%%% 
\begin{table*}[tb]\footnotesize
\begin{center}
\setlength\tabcolsep{2.5pt} 
\begin{tabular}{c|c|c|c|c|ccc|ccc|ccc|ccc|ccc|ccc}
\hline
% header
\multirow{2}{*}{\#} &
\multirow{2}{*}{{\textit{trick}}} & 
\multirow{2}{*}{{\textit{IM}}} & 
\multirow{2}{*}{{\textit{2D}}} & 
\multirow{2}{*}{{\textit{pt}}} & 
\multicolumn{3}{c|}{\textbf{Car AP$_{\text{3D-IoU0.7}}$}} & 
\multicolumn{3}{c|}{{Car AP$_{\text{3D-IoU0.5}}$}} & 
\multicolumn{3}{c|}{\textbf{Ped AP$_{\text{3D-IoU0.5}}$}} &
\multicolumn{3}{c|}{{Ped AP$_{\text{3D-IoU0.25}}$}} &
\multicolumn{3}{c|}{\textbf{Cyclist AP$_{\text{3D-IoU0.5}}$}} &
\multicolumn{3}{c}{{Cyclist AP$_{\text{3D-IoU0.25}}$}} \\
\cline{6-23}
& & & & & Easy & \textbf{Mod} & Hard & Easy & Mod & Hard & Easy & \textbf{Mod} & Hard & Easy & Mod & Hard & Easy & \textbf{Mod} & Hard & Easy & Mod & Hard \\
\hline
\multicolumn{5}{c|}{SECOND~\cite{second} (\textit{teacher})} & 
\textit{91.89} & \textit{81.08} & \textit{78.36} & \textit{99.36} & \textit{94.35} & \textit{93.63} & 
\textit{72.30} & \textit{64.79} & \textit{57.82} & \textit{86.83} & \textit{82.03} & \textit{76.02} & 
\textit{85.17} & \textit{65.81} & \textit{62.00} & \textit{89.31} & \textit{69.41} & \textit{66.07} \\
\hline
\hline
\multicolumn{5}{c|}{DSGN*~\cite{dsgn}} & 
72.31 & 52.29 & 47.12 & 94.73 & 82.33 & 77.17 & 
36.84 & 31.42 & 27.55 & 67.38 & 60.39 & 53.88 & 
35.39 & 23.16 & 22.29 & 46.28 & 30.58 & 29.17 \\
\hline
a. & 
\xmark & & & & 
76.44 & 56.73 & 49.52 & 96.25 & 86.92 & 79.46 & 
24.39 & 19.10 & 16.03 & 73.12 & 60.43 & 52.90 & 
29.12 & 16.27 & 15.14 & 38.81 & 23.36 & 21.49 \\
b. & 
\checkmark & & & & 
82.15 & 62.74 & 57.34 & 98.85 & 90.21 & 84.81 &  
41.18 & 33.72 & 29.19 & 76.19 & 66.04 & 57.98 & 
45.63 & 26.77 & 24.97 & 51.50 & 32.39 & 30.48 \\
\hline
c. & 
\checkmark & \checkmark & & & 
84.99 & 65.67 & 60.16 & 98.84 & 92.47 & 87.32 & 
\textbf{47.82} & \textbf{38.89} & \textbf{33.66} & \textbf{82.78} & \textbf{73.65} & \textbf{64.96} & 
57.54 & 33.70 & 31.23 & 74.39 & 48.44 & 44.93 \\
d. & 
\checkmark & & \checkmark & & 
84.45 & 63.32 & 57.65 & 96.23 & 89.94 & 84.55 & 
40.72 & 34.23 & 29.12 & 76.13 & 66.36 & 58.99 & 
52.65 & 30.26 & 28.14 & 58.59 & 37.96 & 35.42 \\
e. & 
\checkmark & \checkmark & \checkmark & & 
85.26 & 65.64 & 60.12 & \textbf{98.98} & 92.50 & 87.33 & 
43.97 & 36.95 & 31.82 & 77.57 & 69.14 & 62.13 & 
54.59 & 34.07 & 31.42 & 65.19 & 42.96 & 40.28 \\
\hline
f. & 
\checkmark & & & \checkmark & 
81.05 & 63.07 & 56.50 & 98.91 & 92.22 & 87.04 &  
40.29 & 34.13 & 29.31 & 77.60 & 68.54 & 61.42 & 
42.77 & 26.13 & 24.26 & 52.11 & 34.63 & 32.78 \\
g. & 
\checkmark & \checkmark & \checkmark & \checkmark & 
\textbf{86.84} & \textbf{67.71} & \textbf{62.02} & 98.87 & \textbf{93.21} & \textbf{87.97} & 
{45.54} & {37.80} & {32.09} & {81.87} & {72.18} & {64.10} & 
\textbf{60.00} & \textbf{37.31} & \textbf{34.25} & \textbf{79.89} & \textbf{52.17} & \textbf{48.09} \\
\hline
\end{tabular}

\end{center}
\caption{Ablation studies on the KITTI \textit{validation} set. The \textit{tricks} include full-resolution features for stereo volume construction, auxiliary 3D IoU loss, and the uni-modal depth loss as described in Sec.~\ref{sec:losses}. \textit{IM} denotes imitating geometry-aware representations of LiDAR models in Sec.~\ref{sec:distilling-lidar}. \textit{2D} denotes auxiliary direct 2D supervision in Sec.~\ref{sec:aux-2d}. \textit{pt} means ImageNet~\cite{imagenet} pre-trained weights, which helps learn better semantic features. (We only load pre-trained weights for layer \textit{conv1{-}3} due to structure differences). * The results for DSGN~\cite{dsgn} is obtained by evaluating the officially released checkpoints using the new evaluation metrics with 40 recall values for fair comparison.}
\label{table:val-abalation-studies}
\end{table*}

\section{Experiments}
\subsection{Implementation Details}
\label{sec:implementation-details}

\noindent\textbf{Dataset.} We evaluate our method on the challenging KITTI 3D object detection benchmark~\cite{kitti}. 
There are 7,481 training and 7,518 testing stereo image pairs with synchronized LiDAR point clouds in the dataset. 
Following previous papers~\cite{pointrcnn,dsgn}, the training images are split into training set with 3,712 images and validation set with 3,769 images. There are three object classes in KITTI dataset, \textit{car}, \textit{pedestrian}, and \textit{cyclist}, and each class is divided into three difficulty levels, \textit{easy}, \textit{moderate}, and \textit{hard}. 

\noindent\textbf{Evaluation Metric. } We performed detailed evaluation and analysis for 2D, BEV (bird-view) and 3D detection performance. All the performance results are measured using IoU-based criteria to compute mean averaged precision (mAP) over 40 recall values, except for the results in Table~\ref{table:val-results-car}, which are evaluated using the old metric with 11 recall values for fair comparison with previous methods.

\noindent\textbf{Network Structure.} The main structure of our stereo detector follows that of DSGN~\cite{dsgn}, which consists of 2D feature extraction network (left side of Fig.~\ref{fig:framework}(a)), stereo aggregation network (right side of Fig.~\ref{fig:framework}(a)), and BEV feature aggregation network (Fig.~\ref{fig:framework}(b)). As described in Sec.~\ref{sec:losses}, we made several important modifications to \cite{dsgn} to reduce computation cost and memory consumption. 
Compared with \cite{dsgn}, we reduced the numbers of blocks of \textit{conv2} to \textit{conv5} from \{3, 6, 12, 4\} to \{3, 4, 6, 3\}, which is the same as ResNet-34~\cite{resnet}. In addition, we append a small U-Net on the top of the 2D backbone to upsample the SPP feature back into full resolution to provide high-resolution features for stereo matching. 
The number of channels of the stereo aggregation network is halved from 64 to 32, and the 3D hourglass module for the 3D geometry volume $\mathcal{V}_{3d}$ in \cite{dsgn} is removed. For the 3D detection head, we follow the open source implementation of SECOND~\cite{second} in OpenPCDet~\cite{openpcdet} to replace the original FCOS head in \cite{dsgn}, which we found having inferior performance.

For the LiDAR ``teacher'', we employ SECOND~\cite{second} because of its simplicity and similarity with our stereo detector. To make the teacher model and the student model as consistent as possible, we modify the stride of the last sparse convolution layer from 2 to 1, to obtain $1{/}4$ downsampled features to match the feature map size of our stereo detector. Please refer to Sec. 1 of the supplementary materials for detailed network structures of our stereo detector and the LiDAR ``teacher''.

\noindent\textbf{Training Details.} Our stereo detector is trained using AdamW~\cite{adamw} optimizer, with $\beta_1=0.9$, $\beta_2=0.999$. The batch size is fixed to 8. We train the networks with 8 NVIDIA TITAN Xp GPUs, with 1 training sample on each GPU. The model is first trained for 50 epochs using a base learning rate of 0.001, and then for 10 more epochs with a reduced learning rate of 0.0001. The weight decay is set to 0.0001. We apply only horizontal flipping as augmentations. Instead of training individual networks for different classes as \cite{dsgn}, we employ only a single model to train on all three classes of the KITTI dataset. To improve training stability, the weight normalizers for all the loss terms are averaged across all GPUs to avoid unstable gradients. For the hyper-parameters of different losses, $\lambda_{reg}^{L1}{=}0.5$, $\lambda_{reg}^{IoU}{=}1.0$, $\lambda_{dir}{=}0.2$, $\lambda_{2d}{=}1.0$, $\lambda_{im}{=}1.0$. The input size is fixed to $320{\times}1248$, we crop the upper part of input images which does not contain any object to reduce memory consumption. The overall training time is about 12 hours.

For the KITTI dataset, the detection area is set to $[2m, 59.6m]$ for the $Z$ (depth) axis, $[-30m, 30m]$ for the $X$ axis, and $[-1m,3m]$ for the $Y$ axis (in camera coordinate system). The voxel size of $\mathcal{V}_{3d}$ in our stereo model is $(0.2m, 0.2m, 0.2m)$. The input voxel size of the LiDAR detector is set to $(0.05m, 0.1m, 0.05m)$. The spatial resolutions of BEV features of both models are $0.2m{\times}0.2m$. 

\subsection{Main Results}
\label{sec:main-results}

In this section, we provide detailed comparison with state-of-the-art 3D detectors on the KITTI dataset (see Table~\ref{table:test-results-car}, \ref{table:test-results-ped-cyc}, \ref{table:val-abalation-studies}). Our model is only trained with the KITTI dataset, without any pre-training on Scene Flow datasets. 

As shown in Table~\ref{table:val-results-car}, our method surpasses the state-the-of-art method \textit{CG-Stereo} by 8.75\%, 9.24\%, 9.17\% in 3D mAP (IoU=0.7) for \textit{easy}, \textit{moderate} and \textit{hard} difficulty levels of \textit{car} class. Our model even achieves almost the same performance as our LiDAR teacher (see \textit{SECOND (teacher)} item in Table~\ref{table:val-results-car}) based on 0.5-IoU evaluation metrics. The gap between our model and the LiDAR teacher is only 0.2\% mAP for 3D mAP with 0.5 IoU threshold, which proves that stereo-based 3D detection has the potential to replace LiDAR devices as a low-cost solution.

We also submitted our results to the official evaluation benchmark for evaluating our performance on the \textit{test} set of the KITTI dataset. For BEV performance, we surpasses \textit{PLUME-Middle}~\cite{plume} by 10.48\% mAP. For 3D detection performance, we surpasses \textit{CDN}~\cite{cdn} by 10.44\% mAP. Compared with \textit{PL++ (SDN+GDC)}, which is Pseudo-LIDAR++~\cite{pseudo++} using 4-beam LiDAR for refinement, we even achieves much better results. Our method significantly reduces the gap between LiDAR-based methods and stereo-based 3D detection methods. For \textit{pedestrian} and \textit{cyclist} categories, our method exceeds \textit{CG-stereo} by 5.69\% and 5.97\% mAP for 3D detection respectively. Please see Sec. 3 of the supplementary material for visualization results.

\noindent\textbf{Computation cost and memory usage. } Compared with DSGN~\cite{dsgn}, our model has much less layers and number of channels. As shown in Table~\ref{table:time-memory-comparison}, the memory consumption is greatly decreased from 29GB to 10GB for training and 6GB to 4.9GB for testing. For the running time, although we used NVIDIA TITAN Xp, which is expected to be half the speed of NVIDIA V100, our model still decreased half of the running time compared with DSGN.

\subsection{Ablation Study}

In this section, we conduct ablation experiments to validate the contribution of each component of our model on the KITTI validation set. The results are summarized in Table~\ref{table:val-abalation-studies}. The first row \textit{SECOND (teacher)} is the performance of our LiDAR ``teacher''. The model (a.) is our re-implementation of DSGN~\cite{dsgn}, but with less channels and using SECOND~\cite{second} detection head (due to memory limitation and simplicity), which achieves superior performance on cars compared with the original paper, but inferior results for pedestrians and cyclists because we train only a single model for all three classes instead of individual models. We applied a series of important \textit{tricks} to model (a.) to obtain a stronger baseline model (b.), including full-resolution features for stereo volume construction, auxiliary 3D IoU loss for bounding box regression, and the uni-modal depth loss in Eq.~\eqref{equ:loss_depth}. Please refer to Sec. 2 of the supplementary materials for details. By now, we have obtained a stronger baseline model with 61.35\%, 34.11\%, 24.04\% 3D mAPs on the three classes of the KITTI validation set. In the following parts, we will prove the effectiveness of the proposed LiDAR feature imitation strategy in Sec.~\ref{sec:ablation-distill} and the auxiliary direct 2D supervisions in Sec.~\ref{sec:ablation-aux2d}.

\subsubsection{Ablation study of LiDAR feature imitation}
\label{sec:ablation-distill}

\noindent \textbf{Influence of LiDAR feature imitation.} We first study whether imitating LiDAR features can benefit 3D detection and which layer is the optimal one. From Table~\ref{table:abalation-distilling} we can see that each of the three feature layers can provide meaningful geometry-aware guidance, among which $\mathcal{V}_{3d}$ and $\mathcal{\tilde{F}}_{BEV}$ produce the most effective guidance. We also tried different combinations of imitation layers but found no further gain. However, when combining $\mathcal{V}_{3d}$ and $\mathcal{\tilde{F}}_{BEV}$ as the imitation features, we found the model gives more stable results. As a result, we choose $\mathbb{F}_{im}{=}\{\mathcal{V}_{3d}, \mathcal{\tilde{F}}_{BEV}\}$ as our final choice.

\noindent \textbf{Necessity of foreground mask.} 
We then study {the necessity of the foreground mask $\mathcal{M}_{fg}$}. 
If $\mathcal{M}_{fg}$ is removed, the imitation loss would be applied for both foreground and background features with an extreme class imbalance ratio. As expected, after removing $\mathcal{M}_{fg}$, there is almost no improvement in our model (\textit{w/o $\mathcal{M}_{fg}$}) compared with the \textit{baseline} model as shown in Table~\ref{table:abalation-distilling}. 

\noindent \textbf{Imitation weight.} As shown in the last section of Table~\ref{table:abalation-distilling}, we tested different weights $\lambda_{im}$ for the LiDAR imitation loss term and found 1.0 is the optimal choice. 

Therefore, the optimal choice is to imitate foreground features of the 3D volume $\mathcal{V}_{3d}$ and the aggregated BEV features $\mathcal{\tilde{F}}_{BEV}$. In Table~\ref{table:val-abalation-studies}, by comparing model (b.) and (c.), the feature imitation strategy improves 3D detection performance by 2.9\%, 5.2\% and 6.9\% on \textit{car}, \textit{pedestrian} and \textit{cyclist} classes of the KITTI dataset. Similar conclusions can be drawn by comparing model (d.) and (e.).

\subsubsection{The effectiveness of direct 2D supervisions}
\label{sec:ablation-aux2d}

Due to the inefficiency of the 3D supervision for learning semantic features, we appended an auxiliary 2D detection head to provide direct 2D semantic supervisions for the semantic feature $\mathcal{F}_{sem}$. By comparing the model (d.) and the baseline model (b.) in Table~\ref{table:val-abalation-studies}, the direct 2D supervision improves 3D AP of cars from 62.74\% to 63.32\%, pedestrians from 33.72\% to 34.23\%, and cyclists from 26.77\% to 30.26\%. The category with few data, \textit{cyclist}, benefits more from this strategy. 

We also evaluated our semantic features by training 2D detection only, but found poor performance on the KITTI dataset, the 2D APs are only 88\%, 52\% and 36\% for cars, pedestrians and cyclists. We owe the poor results to the lack of data and pre-trained weights. With ImageNet pre-trained weights, the 2D APs are improved to 92\%, 54\% and 41\%. 
To learn more discriminative 2D semantic features, we also performed experiments with ImageNet pre-trained weights to initialize the model. 
The two proposed strategies can consistently improve 3D detection performance with pre-trained weights (see model (f.) and (g.) in Table~\ref{table:val-abalation-studies}).
Our final model (model (g.)) is able to learn both high-quality geometric features and semantic features, which further reduces the gap between LiDAR-based and stereo-based 3D detection algorithms. 
Note that due to the structure differences between our backbone and ResNet-34, we only load pre-trained weights for the first three blocks from \textit{conv1} to \textit{conv3}. For the detailed 2D detection results, please refer to Sec. 2 of the supplementary materials.

\begin{table}[tb]\footnotesize
\begin{center}
\begin{tabular}{c|c|c|c|c}
\hline
\multirow{2}{*}{Method} & Inference & \multirow{2}{*}{GPU} & Memory & AP$_{\text{3D}}$ (\%) \\ 
                         & Time (s)   &  & (GB)        & IoU 0.7 / 0.5  \\
\hline
DSGN \cite{dsgn} & 0.67 & V100 & 29 / 6.0 & 53.95 / 79.10 \\
Ours & 0.35 & TITAN Xp & 10 / 4.9 &  68.47 / 92.85 \\
\hline
\end{tabular}
\end{center}
\caption{Time and memory consumption comparison. The \textit{memory} column shows both training and testing memory usage.}
\label{table:time-memory-comparison}
\end{table}

\begin{table}[tb]\footnotesize
\begin{center}
\begin{tabular}{c|c|c|c}
\hline
Imitation & \multicolumn{3}{c}{Car AP$_{\text{3D}}$} \\ 
\cline{2-4}
Settings & Easy & Moderate & Hard \\
\hline
baseline & 82.15 & 62.74  & 57.34  \\
\hline
$\mathbb{F}_{im}=\{\mathcal{V}_{3d}\}$ & 84.50 & {\bf 65.83} & {\bf 60.35}  \\
$\mathbb{F}_{im}=\{\mathcal{F}_{BEV}\}$ & 84.60 & 65.16 & 58.04 \\
$\mathbb{F}_{im}=\{\mathcal{\tilde{F}}_{BEV}\}$ & 84.83 & 65.10 & 59.42 \\
$\mathbb{F}_{im}=\{\mathcal{V}_{3d},\mathcal{\tilde{F}}_{BEV}\}$ & {\bf 84.99} & 65.67 & 60.16 \\
$\mathbb{F}_{im}=\{\mathcal{V}_{3d},\mathcal{F}_{BEV},\mathcal{\tilde{F}}_{BEV}\}$ & 82.95  & 63.49  & 58.17  \\
\hline
w/o $M_{fg}$ & 81.13 & 62.01 & 54.83 \\
\hline
$\lambda_{im}$=0.2 & 82.61 & 64.15 & 58.23 \\
$\lambda_{im}$=0.5 & 85.14 & 65.44 & 59.85 \\
$\lambda_{im}$=1.0 & {\bf 84.99} & {\bf 65.67} & {\bf 60.16} \\
$\lambda_{im}$=2.0 & 84.47 & 65.11 & 59.84 \\
\hline
\end{tabular}
\end{center}
\caption{Ablation studies for the LiDAR imitation loss. The optimal settings are to imitate $\{\mathcal{V}_{3d},\mathcal{\tilde{F}}_{BEV}\}$ with foreground mask $M_{fg}$, and the loss coefficient is set to $1.0$.}
\label{table:abalation-distilling}
\end{table}

\section{Conclusion}
In this paper, we propose to learn stereo-based 3D detectors under the guidance of high-level geometry-aware LiDAR features and direct semantic supervisions, which successfully improved the geometric and semantic capabilities. Our model surpasses the state-of-the-art algorithms over 10.44\% mAP on the official KITTI 3D detection benchmark, which is closing the gap between stereo-based and LiDAR-based 3D detection algorithms. However, stereo-based 3D detectors still suffer from occlusions, non-textured area and distant objects. By utilizing more advanced ``teachers'' and more robust stereo algorithms, we expect this problem to be solved step by step in the future. 

\section*{Acknowledgement} 
This work is supported in part by the General Research Fund through the Research Grants Council of Hong Kong under Grants (Nos. 14204021, 14208417, 14207319, 14202217, 14203118, 14208619, ), in part by Research Impact Fund Grant No. R5001-18, in part by CUHK Strategic Fund.

{\small
\bibliographystyle{ieee_fullname}
\bibliography{egpaper}

\begin{thebibliography}{10}\itemsep=-1pt

\bibitem{cross_modal_distill_for_lip_reading}
Triantafyllos Afouras, Joon~Son Chung, and Andrew Zisserman.
\newblock Asr is all you need: Cross-modal distillation for lip reading.
\newblock In {\em ICASSP 2020-2020 IEEE International Conference on Acoustics,
  Speech and Signal Processing (ICASSP)}, pages 2143--2147. IEEE, 2020.

\bibitem{aytar2016soundnet}
Yusuf Aytar, Carl Vondrick, and Antonio Torralba.
\newblock Soundnet: Learning sound representations from unlabeled video.
\newblock In D. Lee, M. Sugiyama, U. Luxburg, I. Guyon, and R. Garnett,
  editors, {\em Advances in Neural Information Processing Systems}, volume~29,
  2016.

\bibitem{psmnet}
Jia-Ren Chang and Yong-Sheng Chen.
\newblock Pyramid stereo matching network.
\newblock In {\em CVPR}, pages 5410--5418, 2018.

\bibitem{distill_object_det_chen2017}
Guobin Chen, Wongun Choi, Xiang Yu, Tony Han, and Manmohan Chandraker.
\newblock Learning efficient object detection models with knowledge
  distillation.
\newblock In {\em Proceedings of the 31st International Conference on Neural
  Information Processing Systems}, pages 742--751, 2017.

\bibitem{chen2019_learning_ped_detector_with_distillation}
Rui Chen, Haizhou Ai, Chong Shang, Long Chen, and Zijie Zhuang.
\newblock Learning lightweight pedestrian detector with hierarchical knowledge
  distillation.
\newblock In {\em 2019 IEEE International Conference on Image Processing
  (ICIP)}, pages 1645--1649. IEEE, 2019.

\bibitem{3dop}
Xiaozhi Chen, Kaustav Kundu, Yukun Zhu, Andrew~G Berneshawi, Huimin Ma, Sanja
  Fidler, and Raquel Urtasun.
\newblock 3d object proposals for accurate object class detection.
\newblock In {\em Advances in Neural Information Processing Systems}, pages
  424--432, 2015.

\bibitem{3dop-pami}
Xiaozhi Chen, Kaustav Kundu, Yukun Zhu, Huimin Ma, Sanja Fidler, and Raquel
  Urtasun.
\newblock 3d object proposals using stereo imagery for accurate object class
  detection.
\newblock volume~40, pages 1259--1272. IEEE, 2017.

\bibitem{MV3D}
Xiaozhi Chen, Huimin Ma, Ji Wan, Bo Li, and Tian Xia.
\newblock Multi-view 3d object detection network for autonomous driving.
\newblock In {\em CVPR}, 2017.

\bibitem{chen2018_reality_oriented_adapt_for_sem_seg}
Yuhua Chen, Wen Li, and Luc Van~Gool.
\newblock Road: Reality oriented adaptation for semantic segmentation of urban
  scenes.
\newblock In {\em Proceedings of the IEEE Conference on Computer Vision and
  Pattern Recognition}, pages 7892--7901, 2018.

\bibitem{dsgn}
Yilun Chen, Shu Liu, Xiaoyong Shen, and Jiaya Jia.
\newblock Dsgn: Deep stereo geometry network for 3d object detection.
\newblock {\em Proceedings of the IEEE Conference on Computer Vision and
  Pattern Recognition}, 2020.

\bibitem{nasstereo}
Xuelian Cheng, Yiran Zhong, Mehrtash Harandi, Yuchao Dai, Xiaojun Chang, Tom
  Drummond, Hongdong Li, and Zongyuan Ge.
\newblock Hierarchical neural architecture search for deep stereo matching.
\newblock {\em arXiv preprint arXiv:2010.13501}, 2020.

\bibitem{imagenet}
Jia Deng, Wei Dong, Richard Socher, Li-Jia Li, Kai Li, and Li Fei-Fei.
\newblock Imagenet: A large-scale hierarchical image database.
\newblock In {\em 2009 IEEE conference on computer vision and pattern
  recognition}, pages 248--255. Ieee, 2009.

\bibitem{du2020associate}
Liang Du, Xiaoqing Ye, Xiao Tan, Jianfeng Feng, Zhenbo Xu, Errui Ding, and
  Shilei Wen.
\newblock Associate-3ddet: perceptual-to-conceptual association for 3d point
  cloud object detection.
\newblock In {\em Proceedings of the IEEE/CVF conference on computer vision and
  pattern recognition}, pages 13329--13338, 2020.

\bibitem{feliz2020_squeezed_6dot_detection_distillation}
Heitor Feliz, Walber~M Rodrigues, David Mac{\^e}do, Francisco Sim{\~o}es,
  Adriano~LI Oliveira, Veronica Teichrieb, and Cleber Zanchettin.
\newblock Squeezed deep 6dof object detection using knowledge distillation.
\newblock In {\em 2020 International Joint Conference on Neural Networks
  (IJCNN)}, pages 1--7. IEEE, 2020.

\bibitem{cdn}
Divyansh Garg, Yan Wang, Bharath Hariharan, Mark Campbell, Kilian~Q.
  Weinberger, and Wei-Lun Chao.
\newblock Wasserstein distances for stereo disparity estimation.
\newblock In {\em Advances in Neural Information Processing Systems (NeurIPS)},
  2020.

\bibitem{kitti}
Andreas Geiger, Philip Lenz, and Raquel Urtasun.
\newblock Are we ready for autonomous driving? the kitti vision benchmark
  suite.
\newblock In {\em CVPR}, 2012.

\bibitem{3DSemanticSegmentationWithSubmanifoldSparseConvNet}
Benjamin Graham, Martin Engelcke, and Laurens van~der Maaten.
\newblock 3d semantic segmentation with submanifold sparse convolutional
  networks.
\newblock {\em CVPR}, 2018.

\bibitem{groupwisestereo}
Xiaoyang Guo, Kai Yang, Wukui Yang, Xiaogang Wang, and Hongsheng Li.
\newblock Group-wise correlation stereo network.
\newblock In {\em CVPR}, 2019.

\bibitem{crossmodeldistill_gupta2016}
Saurabh Gupta, Judy Hoffman, and Jitendra Malik.
\newblock Cross modal distillation for supervision transfer.
\newblock In {\em Proceedings of the IEEE conference on computer vision and
  pattern recognition}, pages 2827--2836, 2016.

\bibitem{spp}
Kaiming He, Xiangyu Zhang, Shaoqing Ren, and Jian Sun.
\newblock Spatial pyramid pooling in deep convolutional networks for visual
  recognition.
\newblock {\em IEEE transactions on pattern analysis and machine intelligence},
  37(9):1904--1916, 2015.

\bibitem{resnet}
Kaiming He, Xiangyu Zhang, Shaoqing Ren, and Jian Sun.
\newblock Deep residual learning for image recognition.
\newblock In {\em CVPR}, pages 770--778, 2016.

\bibitem{he2019_knowledge_adapt_for_efficient_sem_seg}
Tong He, Chunhua Shen, Zhi Tian, Dong Gong, Changming Sun, and Youliang Yan.
\newblock Knowledge adaptation for efficient semantic segmentation.
\newblock In {\em Proceedings of the IEEE/CVF Conference on Computer Vision and
  Pattern Recognition}, pages 578--587, 2019.

\bibitem{distill_feature_heo2019comprehensive}
Byeongho Heo, Jeesoo Kim, Sangdoo Yun, Hyojin Park, Nojun Kwak, and Jin~Young
  Choi.
\newblock A comprehensive overhaul of feature distillation.
\newblock In {\em Proceedings of the IEEE/CVF International Conference on
  Computer Vision}, pages 1921--1930, 2019.

\bibitem{distilling_hinton2015}
Geoffrey Hinton, Oriol Vinyals, and Jeff Dean.
\newblock Distilling the knowledge in a neural network.
\newblock {\em arXiv preprint arXiv:1503.02531}, 2015.

\bibitem{distill_featurehuang2017like}
Zehao Huang and Naiyan Wang.
\newblock Like what you like: Knowledge distill via neuron selectivity
  transfer.
\newblock {\em arXiv preprint arXiv:1707.01219}, 2017.

\bibitem{gcnet}
Alex Kendall, Hayk Martirosyan, Saumitro Dasgupta, Peter Henry, Ryan Kennedy,
  Abraham Bachrach, and Adam Bry.
\newblock End-to-end learning of geometry and context for deep stereo
  regression.
\newblock In {\em ICCV}, pages 66--75, 2017.

\bibitem{distill_feature_kim2018paraphrasing}
Jangho Kim, Seonguk Park, and Nojun Kwak.
\newblock Paraphrasing complex network: Network compression via factor
  transfer.
\newblock In S. Bengio, H. Wallach, H. Larochelle, K. Grauman, N. Cesa-Bianchi,
  and R. Garnett, editors, {\em Advances in Neural Information Processing
  Systems}, volume~31. Curran Associates, Inc., 2018.

\bibitem{lang2019pointpillars}
Alex~H Lang, Sourabh Vora, Holger Caesar, Lubing Zhou, Jiong Yang, and Oscar
  Beijbom.
\newblock Pointpillars: Fast encoders for object detection from point clouds.
\newblock In {\em Proceedings of the IEEE/CVF Conference on Computer Vision and
  Pattern Recognition}, pages 12697--12705, 2019.

\bibitem{cgstereo}
Chengyao Li, Jason Ku, and Steven~L Waslander.
\newblock Confidence guided stereo 3d object detection with split depth
  estimation.
\newblock {\em IROS}, 2020.

\bibitem{stereorcnn}
Peiliang Li, Xiaozhi Chen, and Shaojie Shen.
\newblock Stereo r-cnn based 3d object detection for autonomous driving.
\newblock In {\em CVPR}, pages 7644--7652, 2019.

\bibitem{iresnet}
Zhengfa Liang, Yiliu Feng, Yulan Guo, Hengzhu Liu, Wei Chen, Linbo Qiao, Li
  Zhou, and Jianfeng Zhang.
\newblock Learning for disparity estimation through feature constancy.
\newblock In {\em Proceedings of the IEEE Conference on Computer Vision and
  Pattern Recognition}, pages 2811--2820, 2018.

\bibitem{fpn}
Tsung-Yi Lin, Piotr Doll{\'a}r, Ross Girshick, Kaiming He, Bharath Hariharan,
  and Serge Belongie.
\newblock Feature pyramid networks for object detection.
\newblock In {\em Proceedings of the IEEE conference on computer vision and
  pattern recognition}, pages 2117--2125, 2017.

\bibitem{liu2019_structured_distillation_for_sem_seg}
Yifan Liu, Ke Chen, Chris Liu, Zengchang Qin, Zhenbo Luo, and Jingdong Wang.
\newblock Structured knowledge distillation for semantic segmentation.
\newblock In {\em Proceedings of the IEEE/CVF Conference on Computer Vision and
  Pattern Recognition}, pages 2604--2613, 2019.

\bibitem{adamw}
Ilya Loshchilov and Frank Hutter.
\newblock Decoupled weight decay regularization.
\newblock {\em arXiv preprint arXiv:1711.05101}, 2017.

\bibitem{dispnet}
Nikolaus Mayer, Eddy Ilg, Philip Hausser, Philipp Fischer, Daniel Cremers,
  Alexey Dosovitskiy, and Thomas Brox.
\newblock A large dataset to train convolutional networks for disparity,
  optical flow, and scene flow estimation.
\newblock In {\em CVPR}, pages 4040--4048, 2016.

\bibitem{mullapudi2019_online_distillation_for_video_inference}
Ravi~Teja Mullapudi, Steven Chen, Keyi Zhang, Deva Ramanan, and Kayvon
  Fatahalian.
\newblock Online model distillation for efficient video inference.
\newblock In {\em Proceedings of the IEEE/CVF International Conference on
  Computer Vision}, pages 3573--3582, 2019.

\bibitem{crl}
Jiahao Pang, Wenxiu Sun, Jimmy~SJ Ren, Chengxi Yang, and Qiong Yan.
\newblock Cascade residual learning: A two-stage convolutional neural network
  for stereo matching.
\newblock In {\em Proceedings of the IEEE International Conference on Computer
  Vision Workshops}, pages 887--895, 2017.

\bibitem{ida3d}
Wanli Peng, Hao Pan, He Liu, and Yi Sun.
\newblock Ida-3d: Instance-depth-aware 3d object detection from stereo vision
  for autonomous driving.
\newblock In {\em Proceedings of the IEEE/CVF Conference on Computer Vision and
  Pattern Recognition}, pages 13015--13024, 2020.

\bibitem{ocstereo}
Alex~D. Pon, Jason Ku, Chengyao Li, and Steven~L. Waslander.
\newblock Object-centric stereo matching for 3d object detection.
\newblock In {\em 2020 {IEEE} International Conference on Robotics and
  Automation, {ICRA} 2020, Paris, France, May 31 - August 31, 2020}, pages
  8383--8389. {IEEE}, 2020.

\bibitem{qi2018frustum}
Charles~R Qi, Wei Liu, Chenxia Wu, Hao Su, and Leonidas~J Guibas.
\newblock Frustum pointnets for 3d object detection from rgb-d data.
\newblock In {\em Proceedings of the IEEE conference on computer vision and
  pattern recognition}, pages 918--927, 2018.

\bibitem{pseudo_e2e}
Rui Qian, Divyansh Garg, Yan Wang, Yurong You, Serge Belongie, Bharath
  Hariharan, Mark Campbell, Kilian~Q Weinberger, and Wei-Lun Chao.
\newblock End-to-end pseudo-lidar for image-based 3d object detection.
\newblock In {\em Proceedings of the IEEE/CVF Conference on Computer Vision and
  Pattern Recognition}, pages 5881--5890, 2020.

\bibitem{triangulation}
Zengyi Qin, Jinglu Wang, and Yan Lu.
\newblock Triangulation learning network: from monocular to stereo 3d object
  detection.
\newblock {\em IEEE Conference on Computer Vision and Pattern Recognition
  (CVPR)}, 2019.

\bibitem{fasterrcnn}
Shaoqing Ren, Kaiming He, Ross Girshick, and Jian Sun.
\newblock Faster r-cnn: Towards real-time object detection with region proposal
  networks.
\newblock In {\em NIPS}, 2015.

\bibitem{fitnet}
Adriana Romero, Nicolas Ballas, Samira~Ebrahimi Kahou, Antoine Chassang, Carlo
  Gatta, and Yoshua Bengio.
\newblock Fitnets: Hints for thin deep nets.
\newblock In Yoshua Bengio and Yann LeCun, editors, {\em 3rd International
  Conference on Learning Representations, {ICLR} 2015, San Diego, CA, USA, May
  7-9, 2015, Conference Track Proceedings}, 2015.

\bibitem{pvrcnn}
Shaoshuai Shi, Chaoxu Guo, Li Jiang, Zhe Wang, Jianping Shi, Xiaogang Wang, and
  Hongsheng Li.
\newblock Pv-rcnn: Point-voxel feature set abstraction for 3d object detection.
\newblock In {\em Proceedings of the IEEE/CVF Conference on Computer Vision and
  Pattern Recognition}, pages 10529--10538, 2020.

\bibitem{pointrcnn}
Shaoshuai Shi, Xiaogang Wang, and Hongsheng Li.
\newblock Pointrcnn: 3d object proposal generation and detection from point
  cloud.
\newblock In {\em Proceedings of the IEEE/CVF Conference on Computer Vision and
  Pattern Recognition}, pages 770--779, 2019.

\bibitem{shi2020points}
Shaoshuai Shi, Zhe Wang, Jianping Shi, Xiaogang Wang, and Hongsheng Li.
\newblock From points to parts: 3d object detection from point cloud with
  part-aware and part-aggregation network.
\newblock {\em IEEE transactions on pattern analysis and machine intelligence},
  2020.

\bibitem{edgestereo}
Xiao Song, Xu Zhao, Hanwen Hu, and Liangji Fang.
\newblock Edgestereo: A context integrated residual pyramid network for stereo
  matching.
\newblock In {\em Asian Conference on Computer Vision}, 2018.

\bibitem{disprcnn}
Jiaming Sun, Linghao Chen, Yiming Xie, Siyu Zhang, Qinhong Jiang, Xiaowei Zhou,
  and Hujun Bao.
\newblock Disp r-cnn: Stereo 3d object detection via shape prior guided
  instance disparity estimation.
\newblock In {\em CVPR}, 2020.

\bibitem{openpcdet}
OpenPCDet~Development Team.
\newblock Openpcdet: An open-source toolbox for 3d object detection from point
  clouds.
\newblock \url{https://github.com/open-mmlab/OpenPCDet}, 2020.

\bibitem{distilling_fine_grained_detectors_wang}
Tao Wang, Li Yuan, Xiaopeng Zhang, and Jiashi Feng.
\newblock Distilling object detectors with fine-grained feature imitation.
\newblock In {\em Proceedings of the IEEE/CVF Conference on Computer Vision and
  Pattern Recognition}, pages 4933--4942, 2019.

\bibitem{pseudolidar}
Yan Wang, Wei-Lun Chao, Divyansh Garg, Bharath Hariharan, Mark Campbell, and
  Kilian~Q Weinberger.
\newblock Pseudo-lidar from visual depth estimation: Bridging the gap in 3d
  object detection for autonomous driving.
\newblock In {\em CVPR}, pages 8445--8453, 2019.

\bibitem{plume}
Yan Wang, Bin Yang, Rui Hu, Ming Liang, and Raquel Urtasun.
\newblock Plume: Efficient 3d object detection from stereo images.
\newblock {\em arXiv preprint arXiv:2101.06594}, 2021.

\bibitem{mlfstereo}
Bin Xu and Zhenzhong Chen.
\newblock Multi-level fusion based 3d object detection from monocular images.
\newblock In {\em Proceedings of the IEEE conference on computer vision and
  pattern recognition}, pages 2345--2353, 2018.

\bibitem{aanet}
Haofei Xu and Juyong Zhang.
\newblock Aanet: Adaptive aggregation network for efficient stereo matching.
\newblock In {\em Proceedings of the IEEE/CVF Conference on Computer Vision and
  Pattern Recognition}, pages 1959--1968, 2020.

\bibitem{xu2019_training_binary_obj_detector}
Jiaolong Xu, Yiming Nie, Peng Wang, and Antonio~M L{\'o}pez.
\newblock Training a binary weight object detector by knowledge transfer for
  autonomous driving.
\newblock In {\em 2019 International Conference on Robotics and Automation
  (ICRA)}, pages 2379--2384. IEEE, 2019.

\bibitem{zoomnet}
Zhenbo Xu, Wei Zhang, Xiaoqing Ye, Xiao Tan, Wei Yang, Shilei Wen, Errui Ding,
  Ajin Meng, and Liusheng Huang.
\newblock Zoomnet: Part-aware adaptive zooming neural network for 3d object
  detection.
\newblock In {\em Proceedings of the AAAI Conference on Artificial
  Intelligence}, volume~34, pages 12557--12564, 2020.

\bibitem{second}
Yan Yan, Yuxing Mao, and Bo Li.
\newblock Second: Sparsely embedded convolutional detection.
\newblock {\em Sensors}, 18(10):3337, 2018.

\bibitem{yang2018pixor}
Bin Yang, Wenjie Luo, and Raquel Urtasun.
\newblock Pixor: Real-time 3d object detection from point clouds.
\newblock In {\em Proceedings of the IEEE conference on Computer Vision and
  Pattern Recognition}, pages 7652--7660, 2018.

\bibitem{segstereo}
Guorun Yang, Hengshuang Zhao, Jianping Shi, Zhidong Deng, and Jiaya Jia.
\newblock Segstereo: Exploiting semantic information for disparity estimation.
\newblock In {\em ECCV}, pages 636--651, 2018.

\bibitem{3dssd}
Zetong Yang, Yanan Sun, Shu Liu, and Jiaya Jia.
\newblock 3dssd: Point-based 3d single stage object detector.
\newblock In {\em Proceedings of the IEEE/CVF conference on computer vision and
  pattern recognition}, pages 11040--11048, 2020.

\bibitem{std}
Zetong Yang, Yanan Sun, Shu Liu, Xiaoyong Shen, and Jiaya Jia.
\newblock Std: Sparse-to-dense 3d object detector for point cloud.
\newblock In {\em ICCV}, 2019.

\bibitem{yim2017gift}
Junho Yim, Donggyu Joo, Jihoon Bae, and Junmo Kim.
\newblock A gift from knowledge distillation: Fast optimization, network
  minimization and transfer learning.
\newblock In {\em Proceedings of the IEEE Conference on Computer Vision and
  Pattern Recognition}, pages 4133--4141, 2017.

\bibitem{hd3}
Zhichao Yin, Trevor Darrell, and Fisher Yu.
\newblock Hierarchical discrete distribution decomposition for match density
  estimation.
\newblock In {\em Proceedings of the IEEE/CVF Conference on Computer Vision and
  Pattern Recognition}, pages 6044--6053, 2019.

\bibitem{pseudo++}
Yurong You, Yan Wang, Wei-Lun Chao, Divyansh Garg, Geoff Pleiss, Bharath
  Hariharan, Mark Campbell, and Kilian~Q Weinberger.
\newblock Pseudo-lidar++: Accurate depth for 3d object detection in autonomous
  driving.
\newblock In {\em ICLR}, 2020.

\bibitem{ganet}
Feihu Zhang, Victor Prisacariu, Ruigang Yang, and Philip~HS Torr.
\newblock Ga-net: Guided aggregation net for end-to-end stereo matching.
\newblock In {\em CVPR}, pages 185--194, 2019.

\bibitem{atss}
Shifeng Zhang, Cheng Chi, Yongqiang Yao, Zhen Lei, and Stan~Z Li.
\newblock Bridging the gap between anchor-based and anchor-free detection via
  adaptive training sample selection.
\newblock In {\em Proceedings of the IEEE/CVF Conference on Computer Vision and
  Pattern Recognition}, pages 9759--9768, 2020.

\bibitem{unimodal}
Youmin Zhang, Yimin Chen, Xiao Bai, Suihanjin Yu, Kun Yu, Zhiwei Li, and
  Kuiyuan Yang.
\newblock Adaptive unimodal cost volume filtering for deep stereo matching.
\newblock In {\em Proceedings of the AAAI Conference on Artificial
  Intelligence}, volume~34, pages 12926--12934, 2020.

\bibitem{knowledgeaspriors}
Long Zhao, Xi Peng, Yuxiao Chen, Mubbasir Kapadia, and Dimitris~N Metaxas.
\newblock Knowledge as priors: Cross-modal knowledge generalization for
  datasets without superior knowledge.
\newblock In {\em Proceedings of the IEEE/CVF Conference on Computer Vision and
  Pattern Recognition}, pages 6528--6537, 2020.

\bibitem{3diou}
Dingfu Zhou, Jin Fang, Xibin Song, Chenye Guan, Junbo Yin, Yuchao Dai, and
  Ruigang Yang.
\newblock Iou loss for 2d/3d object detection.
\newblock In {\em 2019 International Conference on 3D Vision (3DV)}, pages
  85--94. IEEE, 2019.

\bibitem{zhou2018voxelnet}
Yin Zhou and Oncel Tuzel.
\newblock Voxelnet: End-to-end learning for point cloud based 3d object
  detection.
\newblock In {\em Proceedings of the IEEE Conference on Computer Vision and
  Pattern Recognition}, pages 4490--4499, 2018.

\end{thebibliography}
}



\section*{Supplementary material}

# Supplementary Materials of *LIGA-Stereo: Learning LiDAR Geometry Aware Representations for Stereo-based 3D Detector*

Xiaoyang Guo    Shaoshuai Shi    Xiaogang Wang    Hongsheng Li
CUHK-SenseTime Joint Laboratory, The Chinese University of Hong Kong
{xyguo, ssshi, xgwang, hsli}@ee.cuhk.edu.hk

## 1. More Implementation Details

### 1.1. The Structure of the Stereo Detector

The main structure of our stereo 3D detector follows that of DSGN [2] but with much less memory consumption and lower computation cost. The network structure is described in Table 1 in details.

**2D Feature Extraction.** For the 2D backbone network, we employ a modified version of ResNet-34 [3] with spatial pyramid pooling (SPP) module and feature upsampling. Compared with [2], the numbers of blocks in *conv2-5* are reduced from {3, 6, 12, 4} to {3, 4, 6, 3}. The channels of *conv2-5* are set to {64, 128, 128, 128}. The SPP module is the same as previous implementations [1, 2]. In addition, we append a small U-Net [5] on the top of the 2D backbone to upsample the SPP feature back into full resolution to provide high-resolution features for stereo matching.

**Stereo Aggregation Network.** Although the 2D stereo features $\mathcal{F}_l, \mathcal{F}_r$ are at full image resolution, we still construct the plane sweep volume (the stereo cost volume) at $1/4$ resolution to save memory. The number of base channels of the stereo aggregation network is set to 32, which is half of the number of channels in [2].

**Space Conversion.** The volumetric feature in stereo space $\mathcal{V}_{st}$ is converted into 3D space using Eq. (2) in the paper, which is then filtered by a 3D convolution layer and average pooling layer (along Y dimension) to output the volume in 3D space $\mathcal{V}_{3d}$. The BEV feature is constructed by merging the $y$ dimension and the channel dimension of $\mathcal{V}_{3d}$ and then compressed into 64 channels.

**BEV Aggregation Network.** The structure for the BEV aggregation network follows [2], which is a shallow hourglass-like network.

**3D Detection Head.** For the 3D detection head, we follow the open source implementation of SECOND [7] in OpenPCDet [6]. For each class and each $(x, z)$ location in the BEV space, we create two anchors with fixed average size and rotations of 0 and 90 degrees. The default anchor sizes are $l_a$=3.9, $w_a$=1.6, $h_a$=1.56 for cars, $l_a$=0.8, $w_a$=0.6, $h_a$=1.73 for pedestrians, and $l_a$=1.76, $w_a$=0.6, $h_a$=1.73 for cyclists, and their y coordinates are set to $y_a$={1.78m, 0.6m, 0.6m}, respectively. The training targets are assigned with IoU-based criteria. The matched and unmatched IoU thresholds for the three classes are set to 0.6, 0.5, 0.5 and 0.45, 0.35, 0.35. Anchors with IoU between matched and unmatched thresholds are ignored for training. For each anchor, we output 3-dimensional classification and 7-dimensional regression predictions. The 3D bounding box regression targets are given by the following box encoding functions,

$$\begin{aligned} x_t &= \frac{x_g - x_a}{d_a}, y_t = \frac{y_g - y_a}{h_a}, z_t = \frac{z_g - z_a}{d_a}, \\ w_t &= \log \frac{w_g}{w_a}, l_t = \log \frac{l_g}{l_a}, h_t = \log \frac{h_g}{h_a}, \\ \theta_t &= \theta_g - \theta_a, \end{aligned} \quad (1)$$

where the subscripts $t, a, g$ denote the encoded regression targets, the anchors, and the ground truth. $\theta$ is the yaw direction around the $y$-axis.

The classification loss $\mathcal{L}_{cls}$, the direction classification loss $\mathcal{L}_{dir}$, and the L1 regression loss $\mathcal{L}_{reg}^{L1}$ are the same as SECOND [7], and the auxiliary IoU-based regression loss is defined by,

$$\mathcal{L}_{reg}^{IoU} = 1 - \text{IoU}_{3d}\left(\text{decode}(\delta_p, \xi_a), \xi_g\right) \quad (2)$$

where $\delta_p$ is the regression predictions, $\xi_a$ is the anchor $\{x_a, y_a, z_a, w_a, l_a, h_a, \theta_a\}$, and $\xi_g$ is the assigned ground-truth bounding box $\{x_g, y_g, z_g, w_g, l_g, h_g, \theta_g\}$. Similar to the L1 regression loss, the IoU regression loss is only applied to positive samples. Non-maximum suppression (NMS) is applied to the 3D box predictions for each class separately, with the IoU threshold set to 0.25.

**2D Detection Head.** From the 2D feature extraction part, we have obtained 5-level FPN features *spp1-5* from $\mathcal{F}_{sem}$ with a sequence of stride-2 convolution layers. The strides of these features are {4, 8, 16, 32, 64}. For each level of the features, we apply a 2D detection head with two branches, classification branch and regression branch, to predict 2D bounding boxes. Following ATSS [8], each position is only

| Output | Input | Module Config | #Channel | Size |
|---|---|---|---|---|
| **2D Feature Extraction** | | | | |
| conv1 | $\mathcal{I}_{l/r}$ | Conv (7×7), s=2 | 64 | $H/2 \times W/2$ |
| conv2 | | BasicBlock×3 | 64 | $H/2 \times W/2$ |
| conv3 | | BasicBlock×4, s=2 | 128 | $H/4 \times W/4$ |
| conv4 | | BasicBlock×6, d=2 | 128 | $H/4 \times W/4$ |
| conv5 | | BasicBlock×3, d=4 | 128 | $H/4 \times W/4$ |
| spp1 | | AvgPool (64×64); Conv (1×1); Upsample 64× | 32 | $H/4 \times W/4$ |
| spp2 | | AvgPool (32×32); Conv (1×1); Upsample 32× | 32 | $H/4 \times W/4$ |
| spp3 | | AvgPool (16×16); Conv (1×1); Upsample 16× | 32 | $H/4 \times W/4$ |
| spp4 | | AvgPool (8×8); Conv (1×1); Upsample 8× | 32 | $H/4 \times W/4$ |
| spp | spp1-4, conv3-5 | Concat | 512 | $H/4 \times W/4$ |
| hres1 | conv2 | Conv (1×1) | 64 | $H/2 \times W/2$ |
| hres2 | $\mathcal{I}_{l/r}$ | Conv (1×1) | 32 | $H \times W$ |
| up1 | spp | Conv (3×3); Upsample 2×; Add hres1; ReLU | 64 | $H/2 \times W/2$ |
| up2 | | Conv (3×3); Upsample 2×; Add hres2; ReLU | 32 | $H \times W$ |
| $\mathcal{F}_{l/r}$ | | Conv (3×3)×2 | 32, 32 | $H \times W$ |
| $\mathcal{F}_{sem}$ | spp of $\mathcal{I}_l$ | Conv (3×3)×2 | 128, 32 | $H/4 \times W/4$ |
| fpn_pre | | Conv (1×1) | 64 | $H/4 \times W/4$ |
| fpn0 | | Conv (3×3) | 64 | $H/4 \times W/4$ |
| fpn1 | | Conv (3×3), s=2 | 64 | $H/8 \times W/8$ |
| fpn2 | | Conv (3×3), s=2 | 64 | $H/16 \times W/16$ |
| fpn3 | | Conv (3×3), s=2 | 64 | $H/32 \times W/32$ |
| fpn4 | | Conv (3×3), s=2 | 64 | $H/64 \times W/64$ |
| **Stereo Aggregation Network** | | | | |
| $\mathcal{V}_{st}$ | $\mathcal{F}_l, \mathcal{F}_r$ | Construct Plane Sweep Volume (Eq.(1)) | 64 | $D/4 \times H/4 \times W/4$ |
| st_conv1 | | Conv (3×3×3) | 32 | $D/4 \times H/4 \times W/4$ |
| st_conv2 | | Conv (3×3×3); Add st_conv1 | 32 | $D/4 \times H/4 \times W/4$ |
| st_hg1 | | Conv (3×3×3)×2, s=2 | 64 | $D/8 \times H/8 \times W/8$ |
| st_hg2 | | Conv (3×3×3)×2, s=2 | 64 | $D/16 \times H/16 \times W/16$ |
| st_hg3 | | Deconv (3×3×3); Add st_hg1; ReLU | 64 | $D/8 \times H/8 \times W/8$ |
| $\tilde{\mathcal{V}}_{st}$ | | Deconv (3×3×3); Add st_conv2 | 32 | $D/4 \times H/4 \times W/4$ |
| st_prob | | (3×3×3)×2 | 32, 1 | $D/4 \times H/4 \times W/4$ |
| $\mathcal{P}_{st}$ | | Upsample 4×; Softmax | 1 | $D \times H \times W$ |
| **Stereo Space →3D Space →BEV Space** | | | | |
| $\mathcal{V}_{3d}^{raw}$ | $\mathcal{F}_{sem}, \mathcal{V}_{st}$ | Construct 3D Volume (Eq.(2)) | 64 | $N_x \times N_y \times N_z$ |
| $\mathcal{V}_{3d}$ | | Conv (3×3×1); AvgPool (1×4×1) | 32 | $N_x \times N_y/4 \times N_z$ |
| $\mathcal{F}_{BEV}$ | | Reshape; Conv (3×3) | $32 \times N_y/4$, 64 | $N_x \times N_z$ |
| **BEV Aggregation Network** | | | | |
| bev_hg1 | | Conv (3×3)×2, s=2 | 128 | $N_x/2 \times N_z/2$ |
| bev_hg2 | | Conv (3×3)×2, s=2 | 128 | $N_x/4 \times N_z/4$ |
| bev_hg3 | | Deconv (3×3); Add bev_hg1; ReLU | 128 | $N_x/2 \times N_z/2$ |
| $\tilde{\mathcal{F}}_{BEV}$ | | Deconv (3×3) | 64 | $N_x \times N_z$ |
| **3D Detection Head** | | | | |
| conv_cls | $\tilde{\mathcal{F}}_{BEV}$ | Conv (3×3)×2 | 64 | $N_x \times N_z$ |
| bbox_cls | conv_cls | Conv (3×3) | 6×3 | $N_x \times N_z$ |
| bbox_dir | conv_cls | Conv (1×1) | 6×2 | $N_x \times N_z$ |
| conv_reg | $\tilde{\mathcal{F}}_{BEV}$ | Conv (3×3)×2 | 64 | $N_x \times N_z$ |
| bbox_reg | conv_reg | Conv (3×3) | 6×7 | $N_x \times N_z$ |
| **2D Detection Head** | | | | |
| conv_cls | fpn i (i=0, 1, 2, 3, 4) | Conv (3×3)×4 | 64 | $H/4 \cdot 2^i \times W/4 \cdot 2^i$ |
| bbox_cls | conv_cls | Conv (3×3) | 3 | $H/4 \cdot 2^i \times W/4 \cdot 2^i$ |
| conv_reg | fpn i (i=0, 1, 2, 3, 4) | Conv (3×3)×4 | 64 | $H/4 \cdot 2^i \times W/4 \cdot 2^i$ |
| bbox_reg | conv_reg | Conv (3×3) | 3×4 | $H/4 \cdot 2^i \times W/4 \cdot 2^i$ |
| bbox_centerness | conv_reg | Conv (3×3) | 1 | $H/4 \cdot 2^i \times W/4 \cdot 2^i$ |

Table 1. Detailed network structure of our stereo-based 3D detection network. By default, the convolution layers in the 2D feature extraction module and the 2D head module are followed by batch normalization layers, and the other convolution layers are attached with group normalization layers (the number of groups is set to 32).

| Output | Input | Module Config | #Channel | Size |
|---|---|---|---|---|
| **Sparse 3D Convolution Backbone** | | | | |
| conv1 | input | SpConv (3×3×3) | 16 | $4N_x \times 2N_y \times 4N_z$ |
| conv2 | | SpConv (3×3×3)×3, s=2 | 32 | $2N_x \times N_y \times 2N_z$ |
| conv3 | | SpConv (3×3×3)×3, s=2 | 64 | $N_x \times N_y/2 \times N_z$ |
| conv4 | | SpConv (3×3×3)×3, s=1×2×1 | 64 | $N_x \times N_y/4 \times N_z$ |
| $\mathcal{V}_{3d}$ | | SpConv (1×1×1) | 32 | $N_x \times N_y/4 \times N_z$ |
| $\mathcal{F}_{BEV}$ | | Reshape; Conv (3×3) | $32 \times N_y/4$, 64 | $N_x \times N_z$ |
| **BEV Aggregation Network & 3D Detection Head** | | | | |
| *Same as the stereo detector, see Table 1* | | | | |

Table 2. Detailed network structure of the LiDAR detector.

| IoU | U-Net | Uni-modal | Car AP$_{3D}$ | | |
|---|---|---|---|---|---|
| | | | Easy | Mod | Hard |
| | | | 76.44 | 56.73 | 49.52 |
| ✓ | | | 78.31 | 59.17 | 52.07 |
| ✓ | ✓ | | 78.95 | 59.72 | 54.03 |
| ✓ | ✓ | ✓ | 81.34 | 61.35 | 54.56 |

Table 3. Ablation studies for the *tricks*.

| Supervision | Err. Med. Fg/All (mm) | < 0.2m Fg/All (%) | < 0.4m Fg/All (%) | AP$_{3D}$ (%) |
|---|---|---|---|---|
| Smooth L1 | 0.64 / 0.13 | 61.7 / 37.4 | 40.0 / 22.6 | 78.9 / 59.7 / 54.0 |
| L1 | **0.61** / 0.10 | 55.2 / 32.9 | 35.7 / 20.5 | 78.3 / 61.3 / 54.2 |
| **Hard-assigned** | 0.63 / 0.094 | 50.9 / 29.9 | 32.8 / 18.7 | 80.1 / 61.2 / 54.5 |
| **Gaussian $\sigma$=0.2** | 0.63 / 0.092 | **50.5 / 29.5** | **32.3 / 18.2** | 78.5 / 61.2 / **55.9** |
| Gaussian $\sigma$=0.4 | 0.66 / 0.11 | 55.1 / 32.6 | 35.4 / 19.7 | 78.5 / 61.8 / 54.9 |
| Gaussian $\sigma$=0.8 | 0.70 / 0.13 | 59.5 / 37.1 | 38.4 / 22.0 | 75.7 / 58.8 / 51.9 |
| **Laplacian $\lambda$=0.2** | 0.64 / 0.10 | 52.9 / 31.3 | 33.8 / 19.3 | 80.7 / **62.3** / 55.3 |
| Laplacian $\lambda$=0.4 | 0.66 / 0.11 | 55.5 / 33.4 | 35.6 / 20.3 | 79.0 / 61.4 / 54.6 |
| Laplacian $\lambda$=0.8 | 0.68 / 0.13 | 58.9 / 36.9 | 38.2 / 22.2 | 77.5 / 59.0 / 52.2 |
| **Bilinear (Eq.(6))** | 0.63 / **0.091** | 51.1 / 29.9 | 33.1 / 18.7 | **81.2** / 61.5 / 54.6 |

Table 4. Comparison between different depth losses. *Depth Err. Med.* denotes the average median of depth errors. Foreground (*Fg* in the Table) metrics are evaluated by averaging object-level results, where boxes with less than 5 ground-truth LiDAR points are ignored.

attached with one anchor box. The anchor box sizes for each level are set to $\{32, 64, 128, 256, 512\}$. Please refer to the original paper of ATSS [8] for details.

### 1.2. The Structure of the LiDAR Detector

The main structure of the LiDAR teacher detector follows that of SECOND [7] with minor modifications. We utilize the same BEV aggregation network and 3D detection head as the stereo detector. Detailed network structure is described in Table 2.

## 2. More Ablation Studies

### 2.1. Influence of the *trick* modifications

To improve the performance of the baseline model DSGN [2], we made several important modifications for both training stability and detection performance. The modifications include adding extra IoU-based regression loss, constructing stereo volume using full-resolution 2D feature, and replacing soft-argmin [1] with the uni-modal depth supervision loss. The ablation results are shown in Table 3.

**The effectiveness of the IoU regression Loss.** In previous stereo-based 3D detector implementations, the loss coefficients for regressing locations, orientations, and sizes are usually the same. Although tuning these coefficients has the potential to improve performance, it is trivial and not adaptable. The IoU regression loss [10, 4] instead directly maximizes the 3D IoU between predictions and targets, which can implicitly adapt the regression coefficients. Our results in Table 3 show that the IoU loss can improve about 2.5% mAP for cars.

| 2D Detection Network | pt | AP$_{2D}$ | | |
|---|---|---|---|---|
| | | Car | Ped | Cyc |
| ResNet-34 [3] | ✓ | 88.3 | 66.4 | 49.4 |
| ResNet-34 [3] w/o MaxPool | ✓ | 93.2 | 71.9 | 58.3 |
| Ours (2D only) | | 88.6 | 51.7 | 36.3 |
| Ours (2D only) | ∗ | 91.6 | 54.3 | 41.6 |
| Ours (2D only) | ✓ | 93.2 | 69.2 | 59.2 |
| Ours (Full model) | | 90.8 | 51.9 | 34.3 |
| Ours (Full model) | ∗ | 91.3 | 60.2 | 47.2 |

Table 5. Ablation studies for 2D detection head. ∗ only loads pre-trained weights in *conv1*, *conv2*, and *conv3*.

**Construct stereo volume with high-resolution features.** In PSMNet [1] and DSGN [2], the stereo volume is constructed using left-right image features of $1/4$ size. However, for stereo detection, high-resolution features are essential to improve the depth estimation precision, especially for distant objects. Inspired by the observation, we append an extra small upsampling network (*U-Net*) to construct full-resolution features from SPP features. From Table 3, the *U-Net* improves the 3D detection performance of *moderate* samples by 0.6% mAP and *hard* samples by 2.0% mAP.

**Depth Supervision Loss.** According to [9], indirectly learning cost volume by soft-argmin and smooth-L1 loss is prone to overfitting since the cost volume is under constrained. In comparison, directly minimizing Kullback–Leibler divergence between the predicted distribution and the unimodal distribution centered at true disparities provides stronger constraints to the cost volume, which can learn more robust implicit depth features $\tilde{\mathcal{V}}_{st}$. Since the ground-truth distribution is constant, the KL divergence can be simplified as cross entropy loss with soft targets,

$$\mathcal{L}_{depth} = \frac{1}{N_{gt}} \sum_{u,v} \sum_{w} [-p_w^* \log \mathcal{P}_{st}(u,v,w)], \quad (3)$$

where $p_w^*$ is the ground-truth distribution centered at true disparity $d^*$. Here we investigate several variants of ground-truth distributions, including the bilinearly interpolated distribution (Eq. (6) in the paper), hard-assigned distribution ($p_w$=1 if $d(w)$ is closest to $d^*$), gaussian distribution ($p_w \propto \exp\left(-\frac{1}{2}(\frac{d(w)-d^*}{\sigma})^2\right)$), and laplacian distribution ($p_w \propto \exp\left(-\frac{|d(w)-d^*|}{\lambda}\right)$). The results are shown in Table 4. To evaluate local depth embeddings, instead of using global soft-argmin [1] to parse depth values from depth distributions, we employ local soft-argmin to predict the final depth,

$$\tilde{d}_{u,v} = \sum_{w=k-2}^{k+2} d(w) \cdot \frac{\mathcal{P}_{st}(u,v,w)}{\sum_{w'=k-2}^{k+2} \mathcal{P}_{st}(u,v,w')}, \quad (4)$$

where $k$=argmax$(\mathcal{P}_{st}(u,v,:))$ is the depth index with the

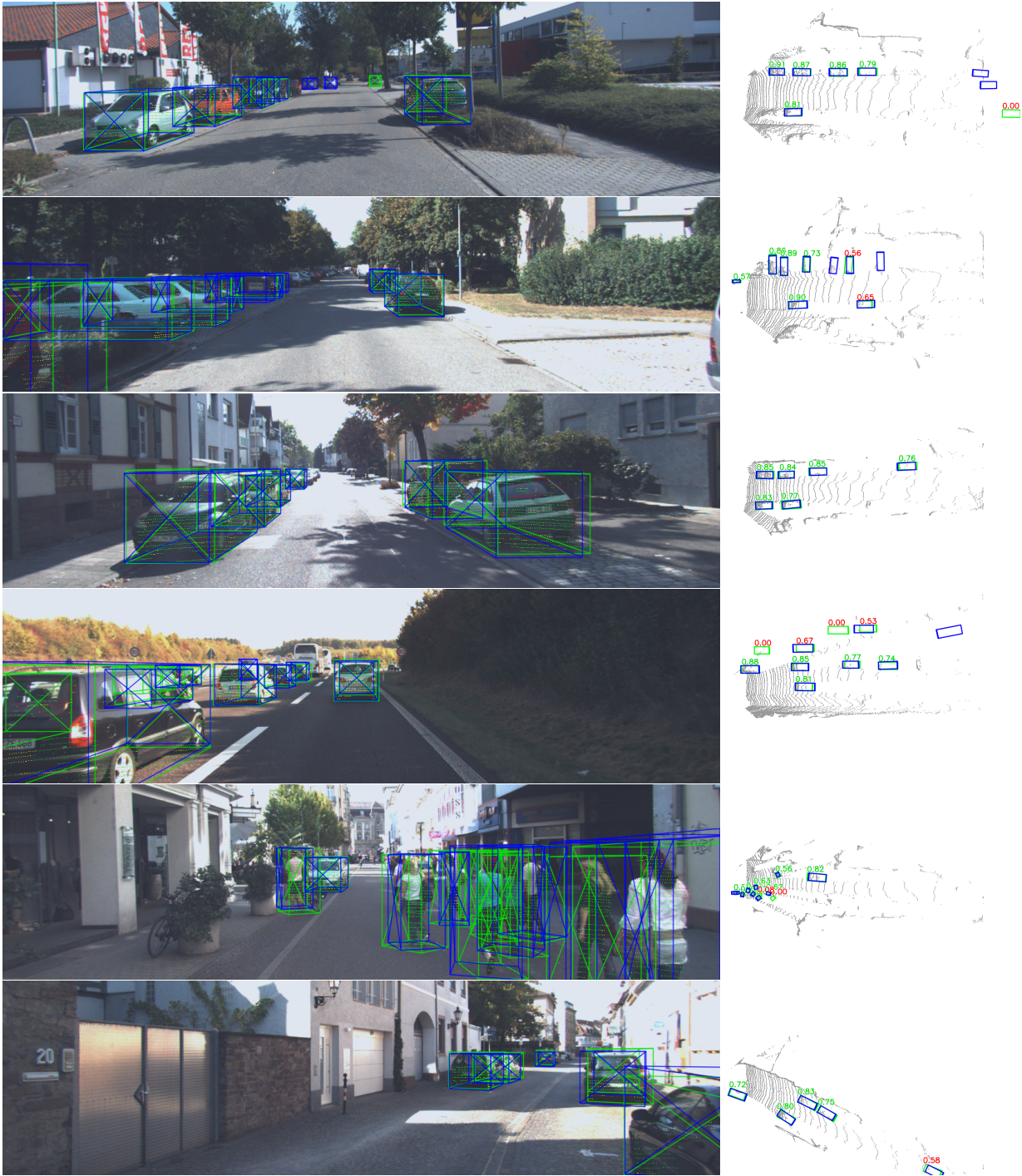

Figure 1. Visualization results of the KITTI *validation* set. The green boxes are ground-truth 3D / BEV bounding boxes. The blue boxes are our predictions. The numbers around the ground-truth BEV boxes are the IoU values of their best predictions. The IoU values will be zero if the corresponding 3D boxes are not detected.

maximum probability. Local soft-argmin can avoid the influence of the probability values that are far from the peak probability, which can be utilized to evaluate the local geometric accuracy of the implicit stereo embeddings $\tilde{\mathcal{V}}_{st}$. Results in Table 4 show that ground-truth distributions $p_w^*$ that are sharper and more concentrated around $d^*$ can give

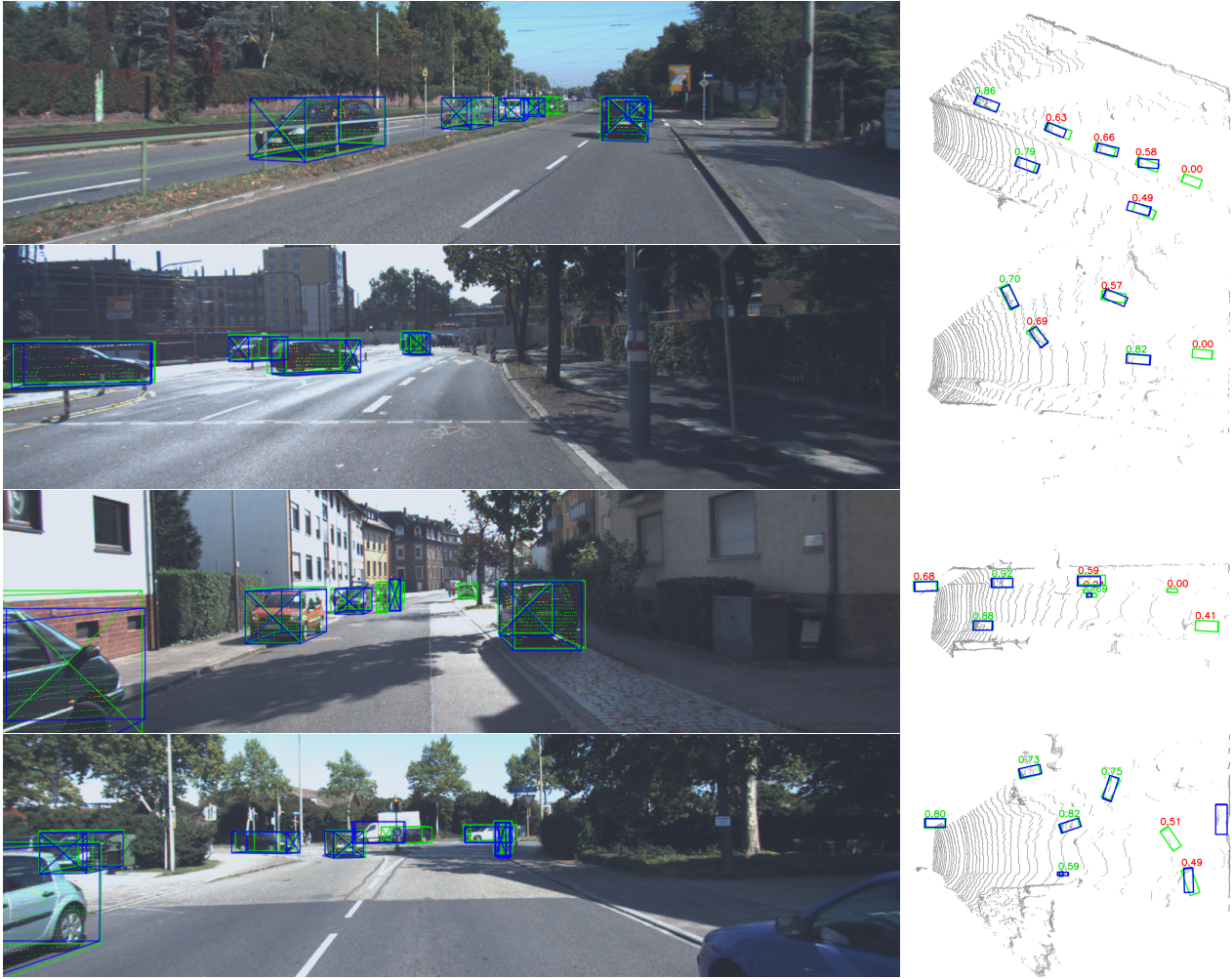

Figure 2. Failure Cases.

better results. The choice of distribution encoding methods is not essential, and hard-assigned distribution can even give better performance than L1 loss. The good choices include hard-assigned distribution, gaussian distribution with $\sigma=0.2$, laplacian distribution with $\lambda=0.2$, and bilinearly interpolated distribution.

## 2.2. 2D Detection Performance

We compare our 2D detection branch with ResNet-34 [3] to confirm that our semantic bottleneck $\mathcal{F}_{sem}$ does not constrain the performance of 2D detection. Since our model does not employ max pooling after *conv1*, we give the results of ResNet-34 without max pooling in the second row of Table 5 for fair comparison. By comparing the results of *ResNet-34 w/o Maxpool* and *Ours (2D only)*, both models achieve similar performance given ImageNet pretrained weights, which proves that the semantic bottleneck does not constrain the 2D detection performance and has the capability of learning good semantic features. By comparing the models without and with ImageNet pretrained weights, we can see pretrained weights are essential for 2D detection due to the limit of training data in the KITTI dataset.

## 3. Visualization Results

Please see the visualization results in Fig. 1. Most of the objects can be detected with high IoU successfully, even for distant objects. We also visualize several failure cases in Fig. 2. Most of the failure cases are caused by occlusions and depth estimation errors. Several predictions give large orientation errors, which we believe can be fixed by incorporating predictions of 2D key-points of bounding boxes in the future.

## References

[1] Jia-Ren Chang and Yong-Sheng Chen. Pyramid stereo matching network. In *CVPR*, pages 5410–5418, 2018. 1,

3

[2] Yilun Chen, Shu Liu, Xiaoyong Shen, and Jiaya Jia. Dsgn: Deep stereo geometry network for 3d object detection. *Proceedings of the IEEE Conference on Computer Vision and Pattern Recognition*, 2020. 1, 3

[3] Kaiming He, Xiangyu Zhang, Shaoqing Ren, and Jian Sun. Deep residual learning for image recognition. In *CVPR*, pages 770–778, 2016. 1, 3, 5

[4] Hamid Rezatofighi, Nathan Tsoi, JunYoung Gwak, Amir Sadeghian, Ian Reid, and Silvio Savarese. Generalized intersection over union: A metric and a loss for bounding box regression. In *Proceedings of the IEEE/CVF Conference on Computer Vision and Pattern Recognition*, pages 658–666, 2019. 3

[5] Olaf Ronneberger, Philipp Fischer, and Thomas Brox. U-net: Convolutional networks for biomedical image segmentation. In *International Conference on Medical image computing and computer-assisted intervention*, pages 234–241. Springer, 2015. 1

[6] OpenPCDet Development Team. Openpcdet: An open-source toolbox for 3d object detection from point clouds. https://github.com/open-mmlab/OpenPCDet, 2020. 1

[7] Yan Yan, Yuxing Mao, and Bo Li. Second: Sparsely embedded convolutional detection. *Sensors*, 18(10):3337, 2018. 1, 3

[8] Shifeng Zhang, Cheng Chi, Yongqiang Yao, Zhen Lei, and Stan Z Li. Bridging the gap between anchor-based and anchor-free detection via adaptive training sample selection. In *Proceedings of the IEEE/CVF Conference on Computer Vision and Pattern Recognition*, pages 9759–9768, 2020. 1, 3

[9] Youmin Zhang, Yimin Chen, Xiao Bai, Suihanjin Yu, Kun Yu, Zhiwei Li, and Kuiyuan Yang. Adaptive unimodal cost volume filtering for deep stereo matching. In *Proceedings of the AAAI Conference on Artificial Intelligence*, volume 34, pages 12926–12934, 2020. 3

[10] Dingfu Zhou, Jin Fang, Xibin Song, Chenye Guan, Junbo Yin, Yuchao Dai, and Ruigang Yang. Iou loss for 2d/3d object detection. In *2019 International Conference on 3D Vision (3DV)*, pages 85–94. IEEE, 2019. 3



\end{document}